\documentclass[sigconf]{acmart}
\AtBeginDocument{%
  }

\copyrightyear{2025}
\acmYear{2025}
\setcopyright{acmlicensed}\acmConference[KDD '25]{Proceedings of the 31st ACM SIGKDD Conference on Knowledge Discovery and Data Mining V.2}{August 3--7, 2025}{Toronto, ON, Canada}
\acmBooktitle{Proceedings of the 31st ACM SIGKDD Conference on Knowledge Discovery and Data Mining V.2 (KDD '25), August 3--7, 2025, Toronto, ON, Canada}
\acmDOI{10.1145/3711896.3737171}
\acmISBN{979-8-4007-1454-2/2025/08}

\usepackage{subfigure}
\usepackage{xspace}
\usepackage{multirow}
\usepackage{xcolor}
\usepackage{amsmath}
\usepackage{amsfonts}
\usepackage{bm}
\usepackage{soul}
\usepackage[ruled]{algorithm2e}
\usepackage{balance}
\usepackage{dsfont}
\usepackage[normalem]{ulem} 

\newcommand{\rev}[1]{{\color{black}{#1}}}

\newcommand{\eat}[1]{}

\newcommand{\boldsubsubsection}[1]{\subsubsection{\textbf{#1}}}

\newcommand{\eg}{\emph{e.g.},\xspace}
\newcommand{\ie}{\emph{i.e.},\xspace}

\newcommand{\etc}{\emph{etc.}\xspace}

\newcommand\figref[1]{Figure~\ref{#1}}

\newcommand\tabref[1]{Table~\ref{#1}}
\newcommand\secref[1]{Section~\ref{#1}}

\newcommand\appref[1]{Appendix~\ref{#1}}
\newcommand{\tabincell}[2]{\begin{tabular}{@{}#1@{}}#2\end{tabular}}  
\newcommand{\model}{{ISTS-PLM}\xspace}
\newcommand{\valstd}[2]{$#1 {\scriptstyle \,\pm\, #2}$}
\newcommand{\valstdb}[2]{$\textbf{#1} {\scriptstyle \,\pm\, #2}$}
\newcommand{\valstdu}[2]{$\underline{#1} {\scriptstyle \,\pm\, #2}$}

\begin{document}


\title{Unleashing The Power of Pre-Trained Language Models for Irregularly Sampled Time Series}

\author{Weijia Zhang}
\authornote{Equal contribution.}
\affiliation{%
  \institution{HKUST(GZ)}
  \city{Guangzhou}
  \country{China}
}
\email{vegazhang3@gmail.com}

\author{Chenlong Yin}
\authornotemark[1]
\affiliation{%
  \institution{HKUST(GZ)}
  \city{Guangzhou}
  \country{China}
}
\email{yinchenlong1@outlook.com}

\author{Hao Liu}
\authornote{Corresponding author.}
\affiliation{%
  \institution{HKUST(GZ) \& HKUST}
  \city{Guangzhou}
  \country{China}
}
\email{liuh@ust.hk}

\author{Hui Xiong}
\authornotemark[2]
\affiliation{%
  \institution{HKUST(GZ) \& HKUST}
  \city{Guangzhou}
  \country{China}
}
\email{xionghui@ust.hk}

\renewcommand{\shortauthors}{Weijia Zhang, Chenlong Yin, Hao Liu, and Hui Xiong}

\begin{abstract}
  Pre-trained Language Models (PLMs), such as ChatGPT, have significantly advanced the field of natural language processing. This progress has inspired a series of innovative studies that explore the adaptation of PLMs to time series analysis, intending to create a unified foundation model that addresses various time series analytical tasks. However, these efforts predominantly focus on Regularly Sampled Time Series (RSTS), neglecting the unique challenges posed by Irregularly Sampled Time Series (ISTS), which are characterized by uneven sampling intervals and prevalent missing data.
  To bridge this gap, this work takes the first step in exploring the potential of PLMs for ISTS analysis. We begin by investigating the effect of various methods for representing ISTS, aiming to maximize the efficacy of PLMs in the analysis. 
  Furthermore, we propose a unified PLM-based framework, named \model, to address diverse ISTS analytical tasks. It integrates novel time-aware and variable-aware PLMs tailored to tackle the intractable intra- and inter-time series modeling in ISTS.
  Finally, extensive experiments on a comprehensive benchmark demonstrate that the \model, utilizing a structured and effective series-based representation for ISTS, consistently achieves state-of-the-art performance across various analytical tasks, such as classification, interpolation, extrapolation, few-shot and zero-shot learning scenarios, spanning scientific domains like healthcare, biomechanics, and climate science. 
  \footnote{Code and datasets are available at: \url{https://github.com/usail-hkust/ISTS-PLM}.}
\end{abstract}

\ccsdesc[500]{Computing methodologies~Machine learning}

\keywords{Irregularly sampled time series; time series foundation models; cross-modal learning}

\maketitle

\section{Introduction}
\label{sec:introduction}
Irregularly Sampled Time Series~(ISTS) are common in diverse domains such as healthcare, biology, climate science, astronomy, physics, and finance~\cite{rubanova2019latent,de2019gru,vio2013irregular,engle1998autoregressive}. 
Although pre-trained foundation models have driven significant progress in natural language processing and computer vision~\cite{zhou2023comprehensive}, their development in time series analysis has been limited by data sparsity and the need for task-specific approaches~\cite{zhou2023one, jin2024time}. 
Recent studies have demonstrated that Pre-trained Language Models (PLMs) possess exceptional abilities in semantic pattern recognition and reasoning across complex sequences~\cite{min2023recent} and proven the universality of PLMs to handle broader data modalities~\cite{zhou2023one, yin2023survey}.
Consequently, we have witnessed that a series of proactive studies explore adapting PLMs for time series analysis~\cite{jin2024position}, highlighting their superiority in generalizability, data efficiency, reasoning ability, multimodal understanding, \etc~\cite{jin2024time,cao2023tempo,pan2024s,wang2025chattime}. 
\rev{The key to their success lies in aligning time series to the semantic space of PLMs, thereby harnessing their powerful sequence modeling capabilities in processing time series.}
However, these studies primarily focus on Regularly Sampled Time Series~(RSTS). 
\rev{The significant challenges of ISTS analysis, stemming from irregular sampling intervals and missing data, raise a critical question: \textit{Are PLMs still effective for ISTS analysis?}}

\begin{figure*}[tb]
  \centering
  \vspace{-2.3mm}
  \includegraphics[width=1.9\columnwidth]{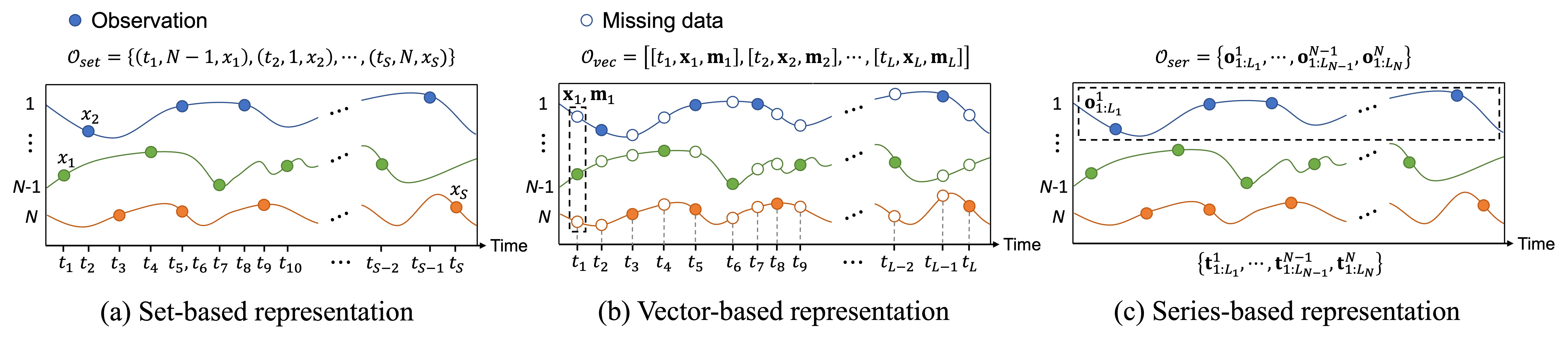}
  \vspace{-3mm}
  \caption{Diverse representation methods for irregularly sampled time series.}
  \vspace{-1mm}
  \label{fig:representation}
\end{figure*}

The modeling and analysis of ISTS differ fundamentally from those of RSTS due to the inherent irregularity and asynchrony among them~\cite{rubanova2019latent,zhangirregular2024}, which further results in diverse representation methods for ISTS tailored to suit different models~\cite{che2018recurrent,horn2020set,shukla2020survey}. 
These distinctive characteristics present the following significant challenges in fully harnessing the capabilities of PLMs for ISTS modeling and analysis:
(1)~\emph{\textbf{Irregularity within ISTS.}}
Unlike applying PLMs to process RSTS, the uneven time intervals between adjacent observations within ISTS disrupt the consistent flow of the series data, making PLMs difficult to identify and capture the real temporal semantics and dependencies. 
For example, positional embeddings~\cite{radford2019language} in PLMs align well with the chronological order of RSTS, where observations occur at fixed intervals. However, it is unsuitable for ISTS since a fixed position may correspond to observations at varying times due to the irregularity of series, which results in inconsistent temporal semantics.
(2)~\emph{\textbf{Asynchrony among ISTS.}} 
While considerable correlations often exist between the time series of different variables, the observations across variables may be significantly misaligned along time dimension due to irregular sampling or missing data~\cite{zhangirregular2024,zhang2021graph}. 
This asynchrony complicates making direct comparisons and discerning correlations between the series, potentially obscuring or distorting the true relationships across variables~\cite{zhang2024irregular_kdd24}. Consequently, this poses a significant challenge in modeling correlations across different time series. 
(3)~\emph{\textbf{Diverse representation methods of ISTS.}}
Unlike RSTS, typically represented as an orderly matrix that comprises a series of vectors containing values of multiple variables, the representation methods for ISTS can vary across different models. Unfortunately, our findings indicate that the commonly used set-based~\cite{horn2020set} and vector-based~\cite{che2018recurrent} representation methods by prior models largely hinder the powerful capabilities of PLMs for ISTS modeling. This imposes a fundamental challenge in identifying a compatible representation method that can stimulate the full potential of PLMs for ISTS analysis.

To bridge the gap, this work takes the first step in exploring the potential of PLMs for ISTS analysis, focusing on the foundational yet often overlooked aspect --- ISTS representation. We investigate the effects of various ISTS representation methods, and reveal that the set-based and vector-based representations, commonly preferred in prior studies, significantly constrain PLMs in analyzing ISTS due to their chaotic data structure. 
This motivates us to introduce a series-based method to represent ISTS in a more structured and effective form. 
Building on this representation, we propose a unified PLM-based framework, \model, which incorporates novel time-aware and variable-aware PLMs tailored to address the challenging intra- and inter-time series modeling in ISTS. By further integrating a learnable input embedding layer and a task output layer, \model is equipped to address diverse ISTS analytical tasks, such as classification, interpolation, and extrapolation.

Our major contributions are summarized as follows:
(1)~This is the first work to explore the potential of PLMs for ISTS analysis, with a focus on the foundational yet often overlooked aspect --- ISTS representation. Our study reveals that the commonly used representation methods limit PLMs’ analytical efficacy, and thus we introduce a more structured and effective series-based representation to maximize the power of PLMs for ISTS.
(2)~We propose time-aware and variable-aware PLMs and integrate them into a unified PLM-based framework to address diverse ISTS analytical tasks, effectively tackling the inherent challenges of ISTS modeling.
(3)~Extensive experiments on a comprehensive benchmark demonstrate that \model consistently achieves state-of-the-art performance across all mainstream ISTS analytical tasks, including classification, interpolation, extrapolation, few-shot and zero-shot learning scenarios, spanning scientific
domains such as healthcare, biomechanics, and climate science.


\begin{figure*}[tb]
  \centering
  \vspace{-3mm}
  \includegraphics[width=1.8\columnwidth]{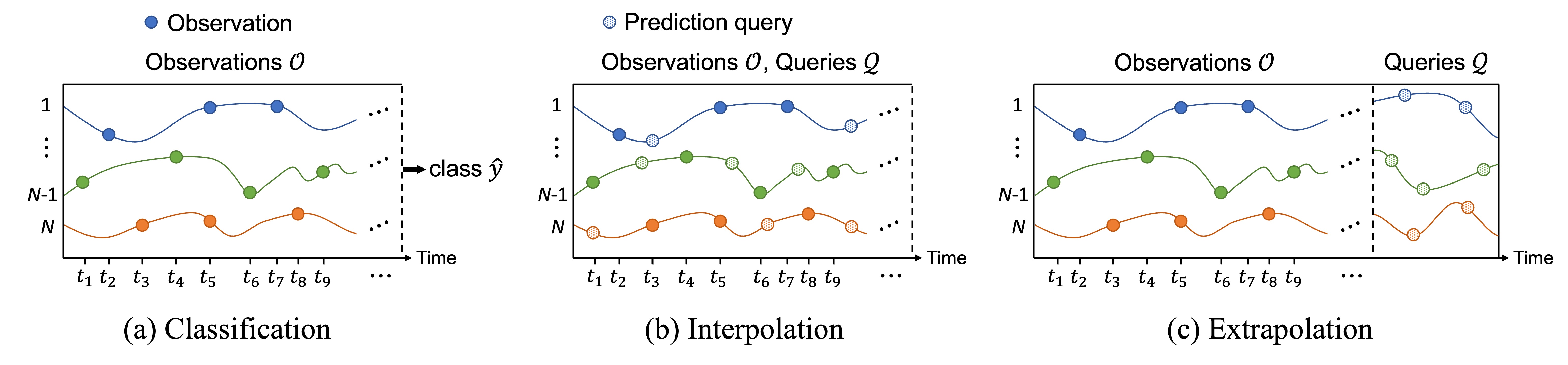}
  \vspace{-3mm}
  \caption{Mainstream analytical tasks for irregularly sampled time series.}
  \label{fig:problem}
  \vspace{-3mm}
\end{figure*}

\section{Related Works}
\label{sec:related}
\subsection{Irregularly Sampled Time Series Analysis}
The primary focus of existing research on ISTS analytical tasks includes classification, interpolation, and extrapolation.
One straightforward method involves converting ISTS into a regularly sampled format~\cite{lipton2016directly}, but this approach often results in significant information loss and missing data issues~\cite{shukla2020multi}. 
Recent studies have shifted towards directly learning from ISTS. 
Specifically, some studies have enhanced RNNs by integrating a time gate~\cite{neil2016phased}, a time decay term~\cite{che2018recurrent}, or memory decomposition mechanism~\cite{baytas2017patient} to adapt the model’s memory updates for ISTS. 
Additionally, inspired by the Transformer's success in processing linguistic sequences and visual data, numerous studies have sought to adapt the Transformer architecture and its attention mechanism for ISTS modeling~\cite{horn2020set, shukla2020multi, zhang2021graph, warpformer2023, li2023time, zhangirregular2024}.
\rev{Another line of studies involve employing neural Ordinary Differential Equations~(ODEs)~\cite{chen2018neural} to capture the continuous dynamics and address the irregularities within ISTS~\cite{rubanova2019latent,de2019gru,bilovs2021neural,schirmer2022modeling,jhin2021ace,jhin2022exit,jhin2023learnable}.} While these works offer a theoretically sound solution, their practical application is usually constrained by high computational costs associated with numerical integration~\cite{bilovs2021neural}. 

Although extensive efforts have been made on ISTS analysis, they primarily focus on addressing a limited range of analytical tasks, with a particular emphasis on classification. 
Furthermore, despite PLMs having demonstrated transformative power in various research areas, such as NLP~\cite{min2023recent}, graph learning~\cite{chen2024exploring}, and even RSTS~\cite{jin2024position}, their potential for ISTS remains unexplored.

\subsection{Pretrained Language Models for Time Series}
We have witnessed that a series of proactive studies explore adapting PLMs for time series analysis not only to enhance task performance but also to facilitate interdisciplinary, interpretative, and interactive analysis~\cite{jin2024position}. 
These studies primarily fall into two categories: prompting-based and alignment-based methods.
Prompting-based methods~\cite{xue2023promptcast, gruver2023large} treat numerical time series as textual data, using existing PLMs to process time series directly. However, the performance is not guaranteed due to the significant differences between time series and text modalities.
Therefore, most recent works focus on alignment-based methods, aiming to align the encoded time series to the semantic space of PLMs, hence harnessing their powerful abilities of semantic pattern recognition and reasoning on processing time series. 
Specifically, model fine-tuning is an effective and the most widely used approach, involving directly tuning partial parameters of PLMs~\cite{zhou2023one,cao2023tempo,liu2024autotimes,pan2024s} or learning additional adapters~\cite{zhou2023one2}.
Moreover, model reprogramming~\cite{jin2024time,sun2023test,wang2025language} aims to directly encode the time series into text representation space that PLMs can understand, thus avoiding tuning PLMs' parameters.

While significant efforts have been made to explore the potential of PLMs for RSTS, harnessing the power of PLMs for ISTS is much more challenging due to its characteristics of irregularity, asynchrony, and diverse representation methods, leaving it largely under-explored.

\section{Preliminary}
\subsection{Representation Methods for ISTS}
Consider that an ISTS $\mathcal{O}$ has $N$ variables, each of which contains a series of observations that are irregularly sampled in varying time intervals. This ISTS can be represented by different methods as illustrated in \figref{fig:representation}.

\textbf{Set-Based Representation.}
Set-based representation method~\cite{horn2020set} views ISTS as a set of observation tuples~$\mathcal{O}_{set} = \{(t_s, n_s, x_s)\}_{s=1}^S$, where $t_s$ is the recorded time, $n_s$ indicates the variable of this observation, $x_s$ denotes the corresponding recorded value, and $S$ represents the total number of observations within the ISTS.

\textbf{Vector-Based Representation.}
Vector-based representation method~\cite{che2018recurrent} has been commonly employed as a standard in current works~\cite{che2018recurrent,shukla2020multi,zhang2021graph,warpformer2023,baytas2017patient,rubanova2019latent,de2019gru,bilovs2021neural,schirmer2022modeling}. This method represents ISTS using three matrix $\mathcal{O}_{vec}=(\mathcal{T}, \mathcal{X}, \mathcal{M})$. $\mathcal{T}=[t_l]_{l=1}^{L} \in \mathbb{R}^L$ represents the unique chronological timestamps of all observations across the ISTS. $\mathcal{X}={[[\tilde{x}^n_l]_{n=1}^N]_{l=1}^{L}}\in \mathbb{R}^{L \times N}$ records the values of variables at these timestamps, with $\tilde{x}^n_l$ representing the observed value of $n$-th variable at time $t_l$, or `NA' if unobserved. $\mathcal{M}={[[m^n_l]_{n=1}^N]_{l=1}^{L}}\in \mathbb{R}^{L \times N}$ is a mask matrix indicating observation status, where $m^n_l=1$ signifies that $\tilde{x}^n_l$ is observed at time $t_l$, and zero otherwise. As a result, the ISTS is represented as a series of vectors $\mathcal{O}_{vec} = \left[t_l, \mathbf{x}_l, \mathbf{m}_l \right]_{l=1}^{L} = \left[t_l, [\tilde{x}^n_l]_{n=1}^N, [m^n_l]_{n=1}^N \right]_{l=1}^{L} \in \mathbb{R}^{L\times (2N+1)}$.

\textbf{Series-Based Representation.}
Series-based representation me-thod represents the time series of each variable separately, and thus leads to $N$ univariant ISTS involving only real observations $\mathcal{O}_{ser} = \{\mathbf{o}^n_{1:L_n}\}_{n=1}^N = \{[(t_i^{n}, x_i^{n})]_{i=1}^{L_n}\}_{n=1}^N$, where $L_n$ represents the number of real observations for $n$-th variable.

\subsection{Problem Definitions}
\figref{fig:problem} showcases the mainstream ISTS analytical tasks studied by existing research works, including classification, interpolation, and extrapolation. 

\emph{\textbf{Problem 1: ISTS Classification.}
Given ISTS observations $\mathcal{O}$, the classification problem is to infer a discrete class $\hat{y}$~(\eg~in-hospital mortality) for the ISTS:
$
\mathcal{F}\left(\mathcal{O}\right) \longrightarrow \hat{y},
$
where $\mathcal{F}(\cdot)$ denotes the classification model we aim to learn.}

\emph{\textbf{Definition 1: Prediction Query.}
A prediction query is denoted as $(t, n)$, indicating a query to predict the recorded value $\hat{x}$ of variable $n$ at time $t$. The queried time may either fall within the observed time window for interpolation or extend beyond it for extrapolation.}

\emph{\textbf{Problem 2: ISTS Interpolation and Extrapolation.}
Given ISTS observations $\mathcal{O}$, and a set of ISTS prediction queries $\mathcal{Q}=\{(t_j,n_j)\}_{j=1}^{|\mathcal{Q}|}$, the problem is to predict recorded values $\hat{\mathcal{X}}=\{\hat{x}_j\}_{j=1}^{|\mathcal{Q}|}$ in correspondence to the prediction queries:
$\mathcal{F}\left(\mathcal{O}, \mathcal{Q} \right) \longrightarrow \hat{\mathcal{X}}.$
}

\vspace{-1mm}
\section{Methodology}
\begin{figure*}[tb]
  \centering
  \includegraphics[width=2\columnwidth]{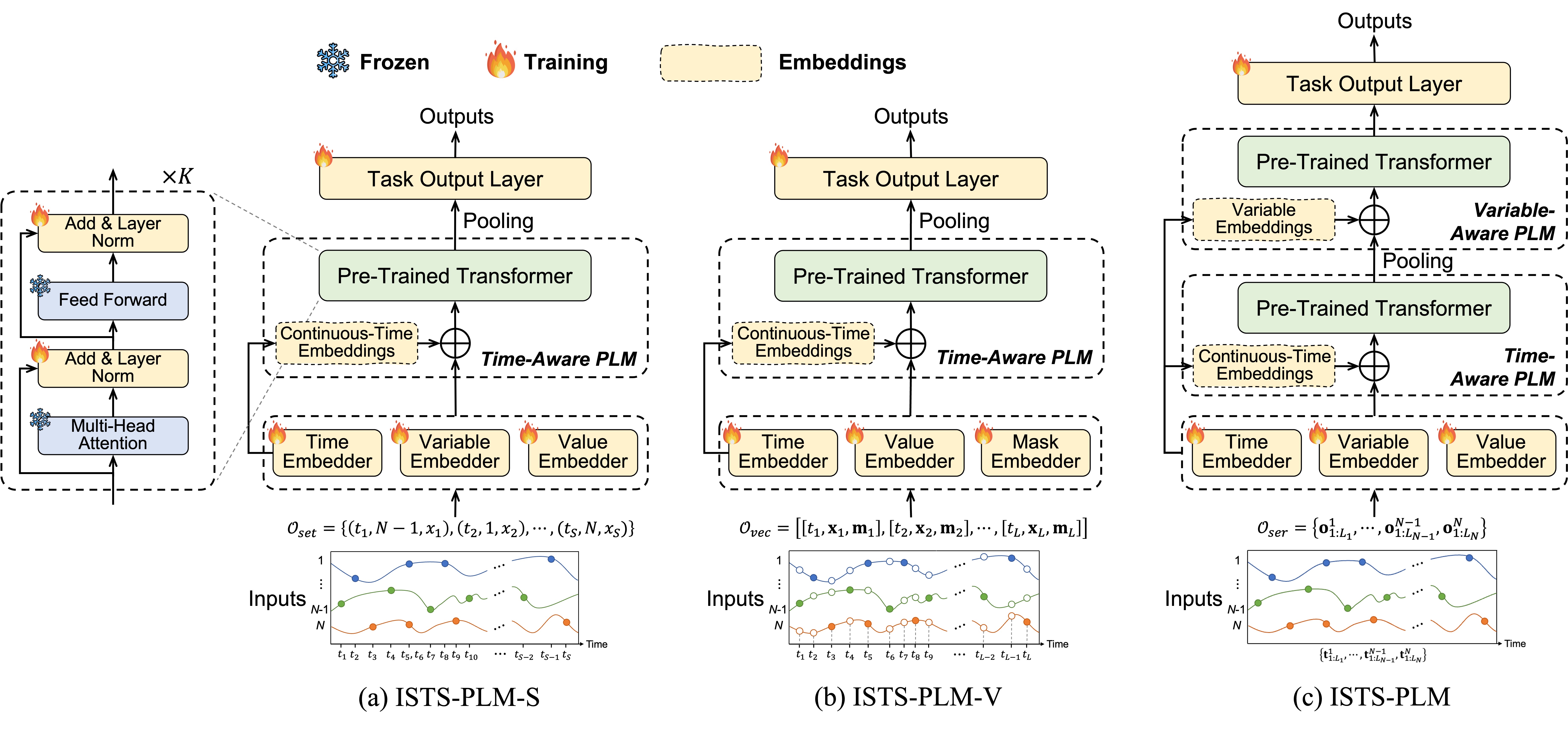}
  \vspace{-3mm}
  \caption{Framework overview of \model for set-based~(\model-S), vector-based~(\model-V), and series-based~(\model) representation methods of ISTS. Given the representation outcome of an ISTS, we first embed it through a trainable input embedding layer to align the embedded ISTS to the semantic space of PLMs. Then the embedded series is fed to a time-aware PLM that replaces its positional embeddings with continuous-time embeddings to discern the irregular dynamics within ISTS. For series-based representation, we further introduce a variable-aware PLM to model variables' correlations within the asynchronous ISTS. All the parameters of PLMs are frozen except layer normalization. Finally, we employ a trainable task output layer to project the output of PLM to address diverse ISTS tasks, such as classification, interpolation, and extrapolation. 
  }
  \label{fig:model}
\end{figure*}
The overview of \model~is illustrated in \figref{fig:model}, where we introduce PLMs for ISTS analysis.
We investigate the effects of set-based, vector-based, and series-based representation methods of ISTS as the inputs for PLMs. 
The unified PLM-based framework, \model, encompasses a trainable input embedding layer, PLM backbone blocks, and a trainable task output layer. Inspired by \cite{zhou2023one}, we freeze all the parameters of the PLMs, except for fine-tuning a few parameters of the layer normalization. Nonetheless, we identify the following key differences from \cite{zhou2023one}: \rev{(1)~\textbf{ISTS Representation}:~We focus on studying diverse representations for ISTS and input model with the outcome of different ISTS representation instead of patching them. (2)~\textbf{Intra-series Modeling}:~To better adapt PLM to model ISTS, we propose time-aware PLM by replacing its positional embeddings with learnable continuous-time embeddings, which empowers PLM to discern the irregular dynamics within ISTS. (3)~\textbf{Inter-series Modeling}:~For series-based representation, we further propose a variable-aware PLM that enables the model to understand and capture the correlations between variables within asynchronous multivariate ISTS, while the work \cite{zhou2023one} only considers intra-series modeling by using a vanilla PLM.}

\vspace{-2mm}
\subsection{Input Embedding}
The input embedding layer aims to align the embedded ISTS to the semantic space of PLMs. Different ISTS representation methods may involve specific subsets of embedders, including time embedder, variable embedder, value embedder, and mask embedder.

\textbf{Time Embedder.} 
To incorporate meaningful temporal information into ISTS modeling, we introduce a time embedder~\cite{shukla2020multi} to encode the continuous-time within ISTS:
\begin{equation}
\boldsymbol{t}[d]= \begin{cases}\omega_{0} \cdot t+\alpha_0, & \text { if } \quad d=0 \\ 
\sin \left(\omega_{d} \cdot t+\alpha_{d}\right), & \text { if } \quad 0<d<D\end{cases}
\end{equation}
\rev{where $d$ represents an index for a vector, $\omega_{d}$ and $\alpha_{d}$ denote learnable parameters, and $D$ is the dimension of continuous-time embedding.}
The linear term captures non-periodic patterns evolving over time, while the periodic terms account for periodicity within the time series, with $\omega_{d}$ and $\alpha_{d}$ indicating the frequency and phase of the sine function, respectively.

\textbf{Variable Embedder.}
This embedder maps the variable $n$ into a $D$ dimensional embedding 
$\boldsymbol{n}$. This can be achieved by utilizing a learnable variable embedding lookup table $\mathbf{V} \in \mathbb{R}^{N \times D}$ and retrieving the corresponding embedding from the lookup table based on the variable indicator, or by using PLM to encode the descriptive text of variable, \etc

\textbf{Value Embedder.}
We adopt a linear embedder layer to encode the recorded values into embeddings. 
For set-based and series-based representations, each recorded value is embedded by:
$\boldsymbol{x} =  x \mathbf{W}_{x}$, where $\mathbf{W}_{x} \in \mathbb{R}^{1 \times D}$ is learnable mapping parameters. 
For vector-based representation, we encode the value vector at each time point into an integrated embedding: $\boldsymbol{x}_l =  \mathbf{x}_l \mathbf{W}_\mathbf{x}$, where $\mathbf{W}_\mathbf{x} \in \mathbb{R}^{N \times D}$. 

\textbf{Mask Embedder.} 
As vector-based representation additionally involves mask terms, we further encode the mask vectors into embeddings. Similar to value embedding, we utilize a linear embedder layer to obtain the mask embedding at each time point: $\boldsymbol{m}_l =  \mathbf{m}_l \mathbf{W}_{m}$, where $\mathbf{W}_{m} \in \mathbb{R}^{N \times D}$ is learnable mapping parameters.

\subsection{PLMs for ISTS Modeling}
This section describes the details of how we adapt PLMs to model ISTS based on set-based, vector-based, and series-based representation methods, respectively.

\boldsubsubsection{PLM for Set-Based Representation.}
Given a set of observation tuples $\mathcal{O}_{set} = \{t_s, n_s, x_s\}_{s=1}^S$, we first sort them in a chronological order. For each tuple, we integrate the embeddings of variable and value: $\boldsymbol{z}_s = \boldsymbol{n}_s + \boldsymbol{x}_s$, obtaining a series of embedded observations $\mathbf{Z}_{set} = \{\boldsymbol{z}_s\}_{s=1}^S \in \mathbb{R}^{S \times D}$, which is then inputted to PLM. 

Due to the irregularity of time series, the same position in PLM might correspond to observations at varying recorded time with completely different temporal semantics.
To empower PLM to seamlessly handle the irregular dynamics within ISTS, we propose time-aware PLM that replaces the positional embeddings of PLM with continuous-time embedding derived from the time embedder. Consequently, the embedded observations will be seamlessly incorporated with temporal information: $\mathbf{Z}'_{set} = \mathbf{Z}_{set} + \mathbf{TE}_{set}$, where $\mathbf{TE}_{set}=\{\boldsymbol{t}_s\}_{s=1}^S \in \mathbb{R}^{S \times D} $ are a series of embeddings of the continuous-times in correspondence to the input observations. 

As the set size $S$ can vary across different ISTS, we summarize PLM's outputs, $\mathbf{H}_{set} \in \mathbb{R}^{S \times D}$, into a fixed dimensional vector, $\mathbf{H}^o_{set} = pool(\mathbf{H}_{set}) \in \mathbb{R}^{D}$, to facilitate the subsequent modeling and analysis, where $pool(\cdot)$ is a size-independent pooling function, such as average, summation, maximum, and attention. 

\boldsubsubsection{PLM for Vector-Based Representation.}
The vector-based representation observations $\mathcal{O}_{vec}$ are first embedded into $\mathbf{Z}_{vec} = \{\boldsymbol{z}_l\}_{l=1}^L \in \mathbb{R}^{L \times D}$, where $\boldsymbol{z}_l = \boldsymbol{x}_l + \boldsymbol{m}_l$. We do not involve variable embedding here because $\boldsymbol{z}_l$ represents an information integration of all variables at time $t_l$, and the value and mask embedders have been variable-aware during this integration. 
Likewise, $\mathbf{Z}_{vec}$ are subsequently processed by a time-aware PLM that seamlessly incorporates the inputs with temporal information, and the output $\mathbf{H}_{vec} \in \mathbb{R}^{L \times D}$ will be summarized into a fixed dimensional vector, $\mathbf{H}^o_{vec}=pool(\mathbf{H}_{vec}) \in \mathbb{R}^{D}$, by a size-independent pooling function.

\boldsubsubsection{PLM for Series-Based Representation.}
PLMs for series-based representation include the processes of intra-time series dependencies modeling and inter-time series correlations modeling. 

\textbf{Intra-Time Series Modeling.} 
This involves modeling each univariate ISTS independently by using a time-aware PLM. Specifically, given $\mathcal{O}_{ser} = \{\mathbf{o}^n_{1:L_n}\}_{n=1}^N = \{[(t_i^{n}, x_i^{n})]_{i=1}^{L_n}\}_{n=1}^N$, a series of observations of $n$-th variable is embedded to $\mathbf{x}^n_{1:L_n} = \{\boldsymbol{x}^n_i\}_{i=1}^{L_n} \in \mathbb{R}^{L_n \times D}$. 
To incorporate variable information, we prepend the variable embedding to the embedding series of each variable: $\mathbf{z}^n_{1:L_n+1} = [\boldsymbol{n},\mathbf{x}^n_{1:L_n}]$. This operation is akin to providing prompts to the inputs of a PLM, enabling it to discern which variable's time series it is analyzing and thus stimulating its in-context learning capability~\cite{lester2021power}.
As a result, $\mathbf{z}^n_{1:L_n+1}$ is inputted to a time-aware PLM for intra-time series dependencies modeling. It initially incorporates continuous-time embeddings to the inputs: $\mathbf{z}^{n'}_{1:L_n+1} = \mathbf{z}^n_{1:L_n+1} + \mathbf{TE}^n_{ser}$, where $\mathbf{TE}^n_{ser} = \left[\boldsymbol{0}, [\boldsymbol{t}^n_i]_{i=1}^{L_n} \right]$ and $\boldsymbol{0}$ represents an all-zero vector. This PLM then processes the inputs through its pre-trained transformer blocks and outputs a series of hidden vectors $\mathbf{h}^n_{1:L_n+1} \in \mathbb{R}^{(L_n+1) \times D}$. 

\textbf{Inter-Time Series Modeling.}
The time series of different variables usually display notable correlations that insights from other variables may offer valuable information and significantly enhance the analysis of each variable~\cite{zhang2021graph,zhangirregular2024,lyu2025autostf}. 
However, the hidden vectors $\{\mathbf{h}^n_{1:L_n + 1}\}_{n=1}^N$ of different variables obtained from the aforementioned intra-time series modeling can be significantly misaligned along time due to their irregularly sampled series, presenting the asynchrony challenge in modeling inter-time series correlations. 
To address this, we similarly summarize the series of hidden vectors into a fixed dimensional vector via the pooling operation: $\mathbf{h}^n_{ser} = pool (\mathbf{h}^n_{1:L_n+1}) \in \mathbb{R}^{D}$, and thus different ISTS will be aligned in a global view.
Once ${\mathbf{H}}_{ser} = \{\mathbf{h}^n_{ser}\}_{n=1}^N \in \mathbb{R}^{N \times D}$ obtained, we employ another variable-aware PLM, which replaces its positional embeddings with the trained variable embeddings: $\mathbf{H}'_{ser} = {\mathbf{H}}_{ser} + \mathbf{VE}_{ser}$, where $\mathbf{VE}_{ser} = \{\boldsymbol{n}\}_{n=1}^N$. 
It facilitates the understanding of variable-specific characteristics and aligns with the position-invariant nature of the variables, thereby enhancing the modeling of inter-time series correlations.
The resulting output produced by this variable-aware PLM is represented as $\mathbf{H}^o_{ser}=\{\mathbf{h}^{o,n}_{ser}\}_{n=1}^N \in \mathbb{R}^{N \times D}$.

\vspace{-2mm}
\subsection{Task Output Projection}
The task output layer aims to project the output of PLMs to address diverse ISTS tasks, \eg~classification, interpolation, extrapolation.

\textbf{Classification.} 
A linear classification layer processes the resulting output of PLMs to infer a class for the ISTS: $\hat{y}=\operatorname{Softmax}(\mathbf{H}^o_{\cdot} \mathbf{W}_c + \mathbf{b}_c) \in \mathbb{R}^{C}$, where $C$ is the number of classes, $\mathbf{W}_c$ and $\mathbf{b}_c \in \mathbb{R}^{C}$ are learnable parameters, and $\mathbf{W}_c \in \mathbb{R}^{D \times C}$ for the outputs of set-based and vector-based representations, $\mathbf{W}_c \in \mathbb{R}^{ND \times C}$ for the flattened output of series-based representation.
The entire learnable parameters of ISTS classification model are trained by optimizing a cross-entropy loss between the inferred class and ground truth label.

\textbf{Interpolation and Extrapolation.}
The output projection of interpolation and extrapolation varies slightly across these representation methods. 
For set-based and vector-based methods, given the resulting output $\mathbf{H}^o_{\cdot}$ of the ISTS and a prediction query $(t, n)$, a prediction layer instantiated by a Multi-Layer
Perception~(MLP) is used to generate the predicted values at time $t$:
$\hat{x} = \operatorname{MLP}([\mathbf{H}^o_{\cdot} \| t])[n]$, \rev{where $\|$ is the concatenation operation.}

For series-based method, we directly utilize the output $\mathbf{h}^{o,n}_{ser} \in \mathbb{R}^D$ of the corresponding variable $n$ to predict its value through a shared prediction layer:
$\hat{x} = \operatorname{MLP}([\mathbf{h}^{o,n}_{ser} \| t])$.

The prediction model is trained by minimizing the Mean Squared Error~(MSE) loss between the prediction and the ground truth: 
$\mathcal{L}_p = \frac{1}{N}\sum_{n=1}^N \frac{1}{Q_n} \sum_{j=1}^{Q_n} \left(\hat{x}^n_j - x^n_j \right)^2,
$
\rev{where $Q_n$ represents the total number of prediction queries in correspondence to $n$-th variable within the predicted windows, which are detailed in \appref{app:datasets}.}

\section{Experiments}
\renewcommand{\arraystretch}{0.9}

\begin{table}[b]
\small
\centering
\vspace{-2mm}
\caption{
Statistics of used irregularly sampled time series datasets for classification, interpolation and extrapolation tasks. 
`Imbalanced' denotes the classes of samples are imbalanced. `Missing ratio' refers to the proportion of missing observations relative to the total number of possible observations in a fully observed dataset.}
\vspace{-3mm}
\label{tab:dataset}
\setlength{\tabcolsep}{0.45mm}{\begin{tabular}{lccccccc}
\toprule
Datasets & \#Samples & \#Variables  & \#Classes & Imbalanced & Missing ratio \\
\hline
P12 & 11,988 & 36  & 2  & True & $88.4\%$ \\
P19 & 38,803 & 34  & 2  & True & $94.9\%$ \\
PAM & 5,333 & 17 & 8  & False & $60.0\%$ \\
\midrule
PhysioNet & 12,000 & 41  & -  & - & $85.7\%$ \\
MIMIC & 23,457 & 96 & - & - & $96.7\%$ \\
Human Activity & 5,400 & 12  & - & - & $75.0\%$ \\
USHCN & 26,736 & 5  & - & - & $95.0\%$  \\
\bottomrule
\end{tabular}}
\end{table}

\begin{table*}[!t]
\small
\centering
\vspace{-3mm}
\caption{Overall performance ($\operatorname{mean} \pm \operatorname{std}$) comparison on irregularly sampled time series \emph{Classification} task. \textbf{Bold} represents the best-performing results and \underline{underline} indicates the second-best results. \model-S, \model-V, \model denote our model with set-based, vector-based, and series-based ISTS representations, respectively.}
\vspace{-3mm}
\label{tab:main_result_classify} 
\begin{tabular}{l|cc|cc|cccc}
\toprule
& \multicolumn{2}{c|}{P12} & \multicolumn{2}{c|}{P19} & \multicolumn{4}{c}{PAM} \\ \cmidrule{2-9}
\multirow{-2}{*}{Method} & AUROC & AUPRC & AUROC & AUPRC & Accuracy & Precision & Recall & F1 score \\ \midrule
Transformer & \valstd{83.3}{0.7} & \valstd{47.9}{3.6} & \valstd{80.7}{3.8} & \valstd{42.7}{7.7} & \valstd{83.5}{1.5} & \valstd{84.8}{1.5} & \valstd{86.0}{1.2} & \valstd{85.0}{1.3} \\
Trans-mean & \valstd{82.6}{2.0} & \valstd{46.3}{4.0} & \valstd{83.7}{1.8} & \valstd{45.8}{3.2} &\valstd{83.7}{2.3} &\valstd{84.9}{2.6} & \valstd{86.4}{2.1}&\valstd{85.1}{2.4}\\
MTGNN & \valstd{74.4}{6.7} & \valstd{35.5}{6.0} & \valstd{81.9}{6.2}  & \valstd{39.9}{8.9} & \valstd{83.4}{1.9} & \valstd{85.2}{1.7} & \valstd{86.1}{1.9} & \valstd{85.9}{2.4}\\
DGM$^2$-O & \valstd{84.4}{1.6} & \valstd{47.3}{3.6} & \valstd{86.7}{3.4} & \valstd{44.7}{11.7} & \valstd{82.4}{2.3} & \valstd{85.2}{1.2} & \valstd{83.9}{2.3} & \valstd{84.3}{1.8}\\
IP-Net& \valstd{82.6}{1.4} &\valstd{47.6}{3.1} & \valstd{84.6}{1.3} & \valstd{38.1}{3.7} &\valstd{74.3}{3.8} & \valstd{75.6}{2.1} & \valstd{77.9}{2.2} & \valstd{76.6}{2.8}\\
GRU-D & \valstd{81.9}{2.1} & \valstd{46.1}{4.7} & \valstd{83.9}{1.7} & \valstd{46.9}{2.1} & \valstd{83.3}{1.6} & \valstd{84.6}{1.2} & \valstd{85.2}{1.6} & \valstd{84.8}{1.2}\\
SeFT & \valstd{73.9}{2.5} & \valstd{31.1}{4.1} & \valstd{81.2}{2.3} & \valstd{41.9}{3.1} & \valstd{67.1}{2.2} & \valstd{70.0}{2.4} & \valstd{68.2}{1.5} & \valstd{68.5}{1.8}\\
mTAND & \valstd{84.2}{0.8} & \valstd{48.2}{3.4} & \valstd{84.4}{1.3} & \valstd{50.6}{2.0} & \valstd{92.9}{0.8} & \valstd{93.8}{0.8} & \valstd{94.0}{0.9} & \valstd{93.8}{0.8}\\ 
Raindrop & \valstd{82.8}{1.7} & \valstd{44.0}{3.0} & \valstd{87.0}{2.3} & \valstd{51.8}{5.5} &\valstd{88.5}{1.5} & \valstd{89.9}{1.5} & \valstd{89.9}{0.6} & \valstd{89.8}{1.0}\\
Warpformer &\valstdu{86.6}{0.8}  &\valstdu{55.5}{3.5} &\valstd{88.1}{2.5} &\valstdu{56.1}{4.4} & \valstd{95.1}{0.8} &\valstd{95.7}{0.8} &\valstd{95.7}{0.9} & \valstd{95.7}{0.8}\\

\cmidrule{1-9}
FPT &\valstd{84.8}{1.1}  &\valstd{50.7}{3.0} &\valstd{87.3 }{2.9} &\valstd{51.6}{3.6} & \valstd{94.0}{1.4} &\valstd{95.3}{0.9} &\valstd{94.7}{1.1} & \valstd{94.9}{1.1}\\
Time-LLM &\valstd{84.4}{1.8} &\valstd{50.2}{1.6} &\valstd{85.1}{2.6} &\valstd{50.1}{3.4} &\valstd{93.4}{1.2} &\valstd{94.2}{1.3} &\valstd{94.7}{1.0} &\valstd{94.4}{1.1} \\
ViTST &\valstd{85.1}{0.8} &\valstd{51.1}{4.1} &\valstdu{89.2}{2.0} &\valstd{53.1}{3.4} & \valstdu{95.8}{1.3} &\valstdu{96.2}{1.3} &\valstdu{96.1}{1.1} & \valstdu{96.5}{1.2} \\

\cmidrule{1-9}
ISTS-PLM-S &\valstd{85.8}{0.9 } &\valstd{52.1 }{4.5 } &\valstd{88.1}{1.4} &\valstd{51.8}{2.2} &\valstd{89.6 }{0.7 } &\valstd{89.6 }{0.7 } &\valstd{91.9 }{1.4 } &\valstd{90.4 }{0.9 }  \\
ISTS-PLM-V &\valstd{85.9 }{1.4 } &\valstd{52.2 }{3.7 } &\valstd{87.5 }{2.4 } &\valstd{53.1 }{4.0 } &\valstd{94.3 }{1.3 } &\valstd{94.5 }{1.4 } &\valstd{95.2 }{1.4 } &\valstd{94.8 }{1.3 } \\
\textbf{ISTS-PLM} 
&\valstdb{87.6}{1.4}  &\valstdb{57.6}{3.3} &\valstdb{89.4}{2.2} &\valstdb{56.9}{5.0}  & \valstdb{96.3}{0.5} &\valstdb{96.9}{1.0} &\valstdb{96.8}{0.4} & \valstdb{96.8}{0.7}\\
\bottomrule
\end{tabular}
\end{table*}

\begin{table*}[!t]
\small
\centering
\caption{Overall performance ($\operatorname{mean} \pm \operatorname{std}$) comparison on irregularly sampled time series \emph{Interpolation} and \emph{Extrapolation} tasks. \textbf{Bold} represents the best-performing results and \underline{underline} indicates the second-best results.}
\vspace{-3mm}
\label{tab:main_result_predict}
\begin{tabular}{c|l|cc|cc|cc}
\toprule
~ & ~ & \multicolumn{2}{c|}{PhysioNet} & \multicolumn{2}{c|}{MIMIC}  & \multicolumn{2}{c}{Human Activity} \\ 
\cmidrule{3-8}
\multirow{-2}{*}{Task} & \multirow{-2}{*}{Method} 
        & MSE$\times10^{-3}$ & MAE$\times10^{-2}$ 
        & MSE$\times10^{-2}$ & MAE$\times10^{-2}$
        & MSE$\times10^{-3}$ & MAE$\times10^{-2}$  \\ 
\midrule
\multirow{15}{*}{\rotatebox{90}{Interpolation}} 
~ & Trans-mean &\valstd{7.76 }{0.18 } &\valstd{4.92}{0.07 }  &\valstd{1.87 }{0.06 } &\valstd{7.63 }{0.18 } &\valstd{3.37 }{0.19 } &\valstd{3.89 }{0.11 }\\
~ & GRU-D &\valstd{6.18}{0.23} &\valstd{4.35}{0.06} &\valstd{2.06 }{0.05 }  &\valstd{7.83 }{0.14 }   &\valstd{2.74}{0.09} &\valstd{3.40}{0.08} \\
~ & SeFT &\valstd{9.46}{0.12} &\valstd{5.51}{0.12} &\valstd{2.12 }{0.02 } &\valstd{8.08 }{0.11 } &\valstd{14.95}{0.03} &\valstd{9.42}{0.02} \\
~ & Raindrop &\valstd{10.65}{0.12} &\valstd{5.80}{0.07} &\valstd{2.31 }{0.04 } &\valstd{8.62 }{0.10 } &\valstd{15.21}{0.12} &\valstd{9.51}{0.05} \\
~ & Warpformer &\valstd{6.37}{0.34} &\valstd{4.43}{0.20} &\valstd{1.93 }{0.06 } &\valstd{7.57 }{0.12 } &\valstd{2.59}{0.15} &\valstd{3.20}{0.11} \\
~ & mTAND &\valstd{5.65}{0.08} &\valstd{4.18}{0.07} &\valstd{1.93 }{0.05 } &\valstd{7.49 }{0.08 } &\valstd{2.07}{0.17} &\valstd{2.74}{0.23} \\
~ & Latent-ODE &\valstd{6.84}{0.34} &\valstd{4.67}{0.14} &\valstd{1.89 }{0.08 } &\valstd{7.67 }{0.47 } &\valstd{3.12}{0.22} &\valstd{3.76}{0.16} \\
~ & Neural Flow &\valstd{6.77 }{0.06 } &\valstd{4.60 }{0.07 } &\valstd{2.18 }{0.11 } &\valstd{8.30 }{0.28 } &\valstd{3.73}{0.06} &\valstd{4.20}{0.05} \\
~ & CRU &\valstd{10.30}{0.10} &\valstd{5.47}{0.05} &\valstd{2.52 }{0.04 } &\valstd{8.87 }{0.03 } &\valstd{7.17}{0.32} &\valstd{6.35}{0.14} \\
~ & t-PatchGNN &\valstdu{4.75}{0.03} &\valstdu{3.59}{0.03}  &\valstdu{1.55 }{0.08 } &\valstdu{6.25 }{0.21 } &\valstdu{1.95 }{0.12 } &\valstdu{2.54 }{0.14 }\\
\cmidrule{2-8}
~ & FPT &\valstd{12.24}{0.07} &\valstd{5.77}{0.12} &\valstd{3.71 }{0.01 } &\valstd{10.76 }{0.09 } &\valstd{2.85}{0.09} &\valstd{3.26}{0.08} \\
~ & Time-LLM &\valstd{12.43 }{0.08 } &\valstd{6.00 }{0.12 } &\valstd{3.63 }{0.05 } &\valstd{10.39 }{0.14 } &\valstd{2.92 }{0.01 } &\valstd{3.34 }{0.03 } \\
\cmidrule{2-8}
~ & \model-S &\valstd{7.74 }{0.35 } &\valstd{5.02 }{0.16 } &\valstd{2.00 }{0.03 } &\valstd{7.71 }{0.08 } &\valstd{4.27 }{0.79 } &\valstd{4.63 }{0.53 } \\
~ & \model-V &\valstd{7.12 }{0.30 } &\valstd{4.58 }{0.08 } &\valstd{1.90 }{0.04 } &\valstd{7.50 }{0.10 } &\valstd{3.03 }{0.22 } &\valstd{3.70 }{0.22 } \\
~ & \textbf{\model} &\valstdb{4.55}{0.08} &\valstdb{3.37}{0.02} &\valstdb{1.47}{0.01} &\valstdb{5.94}{0.01} &\valstdb{1.93}{0.01} &\valstdb{2.52}{0.01} \\

\midrule

\multirow{18}{*}{\rotatebox{90}{Extrapolation}} 
~ & Trans-mean &\valstd{7.45 }{0.26 } &\valstd{4.97 }{0.16 }  &\valstd{1.76 }{0.02 } &\valstd{7.59 }{0.10 } &\valstd{3.65 }{0.11 } &\valstd{4.20 }{0.11 }\\
~ & PatchTST &\valstd{12.00}{0.23} &\valstd{6.02}{0.14}  &\valstd{3.78}{0.03} &\valstd{12.43}{0.10} &\valstd{4.29}{0.14} &\valstd{4.80}{0.09}\\
~ & MTGNN &\valstd{6.26}{0.18} &\valstd{4.46}{0.07} &\valstd{2.71}{0.23} &\valstd{9.55}{0.65} &\valstd{3.03}{0.03} &\valstd{3.53}{0.03} \\
~ & GRU-D & \valstd{5.59}{0.09} & \valstd{4.08}{0.05} &\valstd{1.76}{0.03} &\valstd{7.53}{0.09} &\valstd{2.94}{0.05} & \valstd{3.53}{0.06}\\
~ & SeFT & \valstd{9.22}{0.18} & \valstd{5.40}{0.08} &\valstd{1.87}{0.01} &\valstd{7.84}{0.08} &\valstd{12.20}{0.17} & \valstd{8.43}{0.07}\\
~ & Raindrop & \valstd{9.82}{0.08} & \valstd{5.57}{0.06} &\valstd{1.99}{0.03} &\valstd{8.27}{0.07} &\valstd{14.92}{0.14} & \valstd{9.45}{0.05}\\
~ & Warpformer &\valstd{5.94}{0.35} &\valstd{4.21}{0.12} &\valstd{1.73}{0.04} &\valstd{7.58}{0.13} &\valstd{2.79}{0.04} &\valstd{3.39}{0.03} \\
~ & mTAND &\valstd{6.23}{0.24} &\valstd{4.51}{0.17}  &\valstd{1.85}{0.06} &\valstd{7.73}{0.13} &\valstd{3.22}{0.07} &\valstd{3.81}{0.07}\\
~ & Latent-ODE &\valstd{6.05}{0.57} &\valstd{4.23}{0.17} &\valstd{1.89}{0.19} &\valstd{8.11}{0.52} &\valstd{3.34}{0.11} &\valstd{3.94}{0.12}\\
~ & Neural Flow &\valstd{7.20}{0.07} &\valstd{4.67}{0.04} &\valstd{1.87}{0.05} &\valstd{8.03}{0.19} &\valstd{4.05}{0.13} &\valstd{4.46}{0.09} \\
~ & CRU &\valstd{8.56}{0.26} &\valstd{5.16}{0.09}  &\valstd{1.97}{0.02} &\valstd{7.93}{0.19} &\valstd{6.97}{0.78} &\valstd{6.30}{0.47}\\
~ & t-PatchGNN &\valstdu{4.98}{0.08} &\valstdu{3.72}{0.03}  &\valstdu{1.69}{0.03} &\valstdu{7.22}{0.09} &\valstdu{2.66}{0.03} &\valstdu{3.15}{0.02}\\

\cmidrule{2-8}
~ & FPT &\valstd{10.95}{0.02} &\valstd{5.37}{0.02}  &\valstd{4.00 }{0.03 } &\valstd{12.34 }{0.11 } &\valstd{3.03}{0.09} &\valstd{3.36}{0.10}\\
~ & Time-LLM &\valstd{11.56 }{0.19 } &\valstd{5.80 }{0.25 }  &\valstd{4.41 }{0.01 } &\valstd{13.43 }{0.03 } &\valstd{3.21 }{0.01 } &\valstd{3.53 }{0.02 }\\
\cmidrule{2-8}
~ & \model-S &\valstd{7.62 }{0.20 } &\valstd{4.84 }{0.06 } &\valstd{1.82 }{0.05 } &\valstd{7.78 }{0.07 } &\valstd{5.02 }{1.02 } &\valstd{5.08 }{0.63 } \\
~ & \model-V &\valstd{6.71 }{0.26 } &\valstd{4.52 }{0.11 } &\valstd{1.79 }{0.02 } &\valstd{7.60 }{0.06 } &\valstd{3.46 }{0.16 } &\valstd{4.01 }{0.14 } \\
~ & \textbf{\model} 
&\valstdb{4.92}{0.05} &\valstdb{3.65}{0.04 } &\valstdb{1.64}{0.02} &\valstdb{7.02}{0.14} &\valstdb{2.58}{0.03} &\valstdb{3.12}{0.04}\\
\bottomrule
\end{tabular}
\vspace{-2mm}
\end{table*}

\begin{table*}[tb]
\small
\centering
\vspace{-3mm}
\caption{Few-shot learning task on 10\% data. \textbf{Bold} represents the best-performing results.}
\vspace{-3mm}
\label{tab:few-shot} 
\begin{tabular}{l|cc|cc|cccc}
\toprule
& \multicolumn{2}{c|}{P12} & \multicolumn{2}{c|}{P19} & \multicolumn{4}{c}{PAM} \\ \cmidrule{2-9}
\multirow{-2}{*}{Method} & AUROC & AUPRC & AUROC & AUPRC & Accuracy & Precision & Recall & F1 score \\ 
\midrule
GRU-D &\valstd{79.5 }{2.0 } &\valstd{44.8 }{3.3 } &\valstd{83.3 }{3.0 } &\valstd{33.0 }{3.2 } &\valstd{79.2 }{2.3 } &\valstd{81.4 }{2.5 } &\valstd{80.6 }{1.3 } &\valstd{80.5 }{1.4 } \\
mTAND &\valstd{80.4 }{1.4 } &\valstd{45.2 }{3.5 } &\valstd{76.7 }{1.2 } &\valstd{25.7 }{2.8 } &\valstd{78.7 }{1.6 } &\valstd{81.1 }{0.8 } &\valstd{81.1 }{2.1 } &\valstd{80.6 }{1.4 } \\
Raindrop &\valstd{72.9 }{3.3 } &\valstd{31.6 }{4.9 } &\valstd{75.5 }{3.9 } &\valstd{41.3 }{4.7 } &\valstd{65.6}{4.1 } &\valstd{69.7 }{6.1 } &\valstd{65.3 }{5.6 } &\valstd{65.2 }{6.1 } \\
Warpformer &\valstd{82.2 }{1.3 } &\valstd{45.4 }{3.1 } &\valstd{84.0 }{2.0 } &\valstd{43.7 }{3.8 } &\valstd{84.0 }{1.2 } &\valstd{85.3 }{1.8 } &\valstd{86.4 }{1.6 } &\valstd{85.6 }{1.3 } \\
FPT &\valstd{79.3 }{0.8 } &\valstd{41.6 }{2.6 } &\valstd{81.2 }{2.1 } &\valstd{44.3 }{4.1 } &\valstd{84.2 }{3.0 } &\valstd{86.3 }{2.3 } &\valstd{86.4 }{2.6 } &\valstd{85.9 }{2.7 } \\
Time-LLM &\valstd{79.3 }{2.7 } &\valstd{39.8 }{3.5 } &\valstd{82.6 }{2.7 } &\valstd{39.5 }{4.4 } &\valstd{83.4 }{1.7 } &\valstd{84.5 }{1.3 } &\valstd{84.5 }{2.0 } &\valstd{85.0 }{1.5 } \\
ViTST &\valstd{69.6 }{2.1 } &\valstd{29.5 }{2.3 } &\valstd{71.3 }{1.6 } &\valstd{33.2 }{1.4 } &\valstd{80.3 }{3.2 } &\valstd{85.1 }{2.1 } &\valstd{81.8 }{3.2 } &\valstd{82.6 }{3.3 } \\
\cmidrule{1-9}
\textbf{\model} &\valstdb{83.8}{1.7}  &\valstdb{49.6}{3.9} &\valstdb{85.3}{2.2} &\valstdb{45.9}{5.0}  & \valstdb{85.7}{1.3} &\valstdb{87.8}{1.0} &\valstdb{87.7}{1.6} & \valstdb{87.5}{1.1}\\
\bottomrule
\end{tabular}
\vspace{-3mm}
\end{table*}

\vspace{-1mm}
\subsection{Experimental Setup}
\label{sec:setup}
To demonstrate the effectiveness of \model, we conduct extensive experiments across mainstream ISTS analytical tasks, including classification, interpolation, and extrapolation. 
More experiments studying the effect of the number of used PLMs' layers and PLMs configurations are provided in \ref{app:para_sens}, \ref{app:diff_plms} in Appendix.

\vspace{-1mm}
\boldsubsubsection{Datasets}
For ISTS classification task, we refer to previous works~\cite{li2023time,zhang2021graph} that introduce healthcare datasets P12~\cite{physionet}, P19~\cite{reyna2019early} and biomechanics dataset PAM~\cite{reiss2012introducing} for a thorough evaluation. 
We follow the experimental settings of P12, P19, and PAM from ViTST~\cite{li2023time}, where each dataset is randomly partitioned into training, validation, and test sets with 8:1:1 proportion. 
Each experiment is performed with five different data partitions and reports the mean and standard deviation of results. 
The indices of these partitions are kept consistent across all methods compared. 

For ISTS interpolation and extrapolation tasks, referring to t-PatchGNN~\cite{zhangirregular2024}, we utilize four datasets: PhysioNet~\cite{physionet}, MIMIC~\cite{mimic3}, and Human Activity~\cite{misc_localization_data_for_person_activity_196}, USHCN~\cite{USHCN} from the domains of healthcare, biomechanics, and climate science. 
Consistently, we randomly divide all the ISTS samples within each dataset into training, validation, and test sets, maintaining a proportional split of 6:2:2, and adopt the min-max normalization (z-score normalization for USHCN) to normalize the original observation values. 
To mitigate randomness, we run each experiment with five different random seeds and report the mean and standard deviation of the results. 
The statistics of used datasets are displayed in \tabref{tab:dataset}, and more details are provided in \appref{app:datasets}.

\boldsubsubsection{Metrics}
For classification task, following prior research~\cite{li2023time, zhang2021graph}, we utilize Area Under the Receiver Operating Characteristic Curve~(AUROC) and Area Under the Precision-Recall Curve~(AUPRC) for the performance evaluation of imbalanced datasets P12 and P19, and use Accuracy, Precision, Recall, and F1 score to evaluate balanced dataset PAM. Higher is better for all the above metrics.

Referring to previous work~\cite{zhangirregular2024}, we introduce both Mean Square Error~(MSE) and Mean Absolute Error~(MAE) to evaluate the prediction performance for interpolation and extrapolation tasks. Lower is better for MSE and MAE.

\boldsubsubsection{Baselines}
To evaluate the performance in ISTS classification task, we incorporate the following baseline models for a fair comparison, including vanilla Transformer~\cite{vaswani2017attention}, \rev{Trans-mean~(Transformer with mean imputation for missing values)}; (sparse) multivariate time series analysis models: MTGNN~\cite{wu2020connecting}, DGM$^2$-O~\cite{wu2021dynamic}; ISTS classification models:
IP-Net~\cite{shukla2018interpolation}, GRU-D~\cite{che2018recurrent}, SeFT~\cite{horn2020set},
mTAND~\cite{shukla2020multi}, Raindrop~\cite{zhang2021graph}, Warpformer~\cite{warpformer2023}, and pre-trained vision transformers-based model ViTST~\cite{li2023time};
as well as PLM-based models designed for RSTS analysis: 
FPT~\cite{zhou2023one}, Time-LLM~\cite{jin2024time}. 
All these models are trained for 20 epochs, and the model's parameters achieving the highest AUROC on the validation set are selected for testing~\cite{zhang2021graph,li2023time}. 

For ISTS interpolation and extrapolation tasks, except adapting the representative baselines above to these two tasks, we further incorporate several models tailored for the ISTS prediction tasks, including Latent-ODE~\cite{rubanova2019latent}, Neural Flow~\cite{bilovs2021neural}, CRU~\cite{schirmer2022modeling}, and t-PatchGNN~\cite{zhangirregular2024}.
For both tasks, early stopping is applied to all models if the validation loss doesn’t decrease over 10 epochs. 
More details of these baselines are provided in \appref{app:baseline}.

\boldsubsubsection{Implementation Details}
\label{app:implement}
While most of the experiments are conducted on a Linux server with a 20-core Intel(R) Xeon(R) Platinum 8255C CPU @ 2.50GHz and an NVIDIA Tesla V100 GPU, the interpolation and extrapolation experiments of PhysioNet and MIMIC datasets are performed on a 8-core Intel(R) Xeon(R) Platinum 8358P CPU @ 2.60GHz and an NVIDIA A800 GPU.
We use the first 6 layers~(out of 12) of GPT-2 (124M)\footnote{\url{https://huggingface.co/openai-community/gpt2}} as time-aware PLM and the first 6 layers~(out of 12) of BERT (110M)\footnote{\url{https://huggingface.co/google-bert/bert-base-uncased}} as variable-aware PLM in \model for classification tasks and Human Activity's interpolation and extrapolation. We use the first 3 layers of PLMs for tasks on PhysioNet and MIMIC, and use the first layer for USHCN. 
The embedding dimension $D$ is 768. 
We freeze all parameters of the PLMs, except for fine-tuning only a few parameters of the layer normalization.
For simplification, we learn a variable embedding lookup table as the variable embedder and use an average pooling function. 
The model is trained using the Adam optimizer with a learning rate $1\times10^{-3}$ for classification and $5\times10^{-4}$ for the other tasks.
\model~employs consistent dataset splitting and validation strategies as the baseline models to ensure fair comparison.


\vspace{-1mm}
\subsection{Main Results}
\tabref{tab:main_result_classify}, \tabref{tab:main_result_predict}, and \tabref{table:ushcn_performance} present the performance comparison for ISTS classification, interpolation and extrapolation tasks. Our \model (using series-based representation) outperforms all other baselines, including the state-of-the-art~(SOTA) cross-domain adaptation-based methods: FPT, Time-LLM, and ViTST, across all tasks and datasets~(except MSE on USHCN), demonstrating \model's universal superiority for ISTS analysis. However, it is non-trivial to obtain this level of performance. We observe that \model with typical set-based~(ISTS-PLM-S) and vector-based~(ISTS-PLM-V) representations often yield sub-optimal results. They perform even worse on interpolation and extrapolation tasks, which require each variable to be more meticulously and distinctly analyzed. We provide further analysis on different ISTS representations of our model in \secref{sec:diff_reps}. 

\rev{Additionally, although we utilize the same series-based representation of ISTS as inputs for the PLM-based models FPT and Time-LLM, they exhibit unsatisfactory performance, particularly in interpolation and extrapolation tasks. This is mainly because they handle each variable's time series independently and fail to model inter-series correlations, which has been proven to be crucial for these prediction tasks.}

\label{sec:ablation}
\begin{table*}[tb]
\small
\centering
\vspace{-1mm}
\caption{Model ablation results on \emph{Classification} task. \textbf{Bold} represents the best-performing results.}
\vspace{-3mm}
\label{tab:ablation_classification} 
\begin{tabular}{l|cc|cc|cccc}
\toprule
& \multicolumn{2}{c|}{P12} & \multicolumn{2}{c|}{P19} & \multicolumn{4}{c}{PAM} \\ \cmidrule{2-9}
\multirow{-2}{*}{Method} & AUROC & AUPRC & AUROC & AUPRC & Accuracy & Precision & Recall & F1 score \\ 
\midrule
\textbf{ISTS-PLM} &\valstdb{87.6}{1.4}  &\valstdb{57.6}{3.3} &\valstdb{89.4}{2.2} &\valstdb{56.9}{5.0}  & \valstdb{96.3}{0.5} &\valstd{96.9}{1.0} &\valstdb{96.8}{0.4} & \valstdb{96.8}{0.7}\\
\cmidrule{1-9}
Random &\valstd{53.3  }{3.7 } &\valstd{16.6 }{3.1 } &\valstd{60.9 }{4.7 } &\valstd{6.3 }{1.2 } &\valstd{21.4 }{2.0 } &\valstd{2.7 }{0.5 } &\valstd{12.6 }{0.2 } &\valstd{4.4 }{0.3 } \\
w/o TA-PLM &\valstd{86.4 }{1.0 } &\valstd{54.4 }{4.1 } &\valstd{89.1 }{2.4 } &\valstd{54.2 }{4.1 } &\valstd{95.0 }{0.6 } &\valstd{95.7 }{0.5 } &\valstd{95.7 }{0.5 } &\valstd{95.6 }{0.5 } \\
w/o VA-PLM &\valstd{87.0 }{1.2 } &\valstd{55.4 }{2.9 } &\valstd{87.6 }{2.7 } &\valstd{51.5 }{4.1 } &\valstd{91.5 }{1.9 } &\valstd{92.4 }{2.3 } &\valstd{93.1 }{1.5 } &\valstd{92.3 }{2.4 } \\
w/o TE &\valstd{87.2 }{0.9 } &\valstd{57.2 }{3.5 } &\valstd{88.3 }{2.6 } &\valstd{53.3 }{3.4 } &\valstd{96.1 }{0.4 } &\valstdb{97.0}{0.4} &\valstd{96.4 }{0.6 } &\valstd{96.6 }{0.3 } \\
w/o VE &\valstdb{87.6}{1.6} &\valstd{57.2 }{3.6 } &\valstd{88.6 }{2.3 } &\valstd{55.4 }{4.0 } &\valstd{95.5 }{1.3 } &\valstd{96.2 }{1.3 } &\valstd{96.0 }{0.6 } &\valstd{96.1 }{1.0 } \\
\bottomrule
\end{tabular}
\vspace{-1mm}
\end{table*}

\begin{table*}[tb]
\small
\centering
\caption{Model ablation results on \emph{Interpolation} and \emph{Extrapolation} tasks. \textbf{Bold} represents the best-performing results.}
\vspace{-3mm}
\label{tab:ablation_prediction}
\begin{tabular}{c|l|cc|cc|cc}
\toprule
~ & ~ & \multicolumn{2}{c|}{PhysioNet} & \multicolumn{2}{c|}{MIMIC} & \multicolumn{2}{c}{Human Activity} \\ 
\cmidrule{3-8}
\multirow{-2}{*}{Task} & \multirow{-2}{*}{Method} 
        & MSE$\times10^{-3}$ & MAE$\times10^{-2}$ 
        & MSE$\times10^{-2}$ & MAE$\times10^{-2}$ 
        & MSE$\times10^{-3}$ & MAE$\times10^{-2}$  \\ 
\midrule
\multirow{7}{*}{\rotatebox{90}{Interpolation}} 
~ & \textbf{\model} 
&\valstdb{4.55}{0.08} &\valstdb{3.37}{0.02} &\valstdb{1.47}{0.01} &\valstdb{5.94}{0.01} &\valstdb{1.93}{0.01} &\valstdb{2.52}{0.01}\\
\cmidrule{2-8}
~ & Random &\valstd{68.50 }{0.01 } &\valstd{20.38 }{0.03 } &\valstd{7.01 }{0.01 } &\valstd{21.32 }{0.11 } &\valstd{16.83 }{0.12 } &\valstd{10.36 }{0.05 }\\
~ & w/o TA-PLM &\valstd{5.47 }{0.18 } &\valstd{3.78 }{0.08 }  &\valstd{1.49 }{0.02 } &\valstd{6.05 }{0.08 } &\valstd{2.85 }{0.01 } &\valstd{3.29 }{0.01 } \\
~ & w/o VA-PLM &\valstd{4.94 }{0.13 } &\valstd{3.52 }{0.10 } & \valstd{1.88 }{0.04 } &\valstd{7.48 }{0.15 } &\valstd{2.53 }{0.54 } &\valstd{3.02 }{0.44 }\\
~ & w/o TE &\valstd{4.85 }{0.11 } &\valstd{3.52 }{0.11 } &\valstdb{1.47}{0.01} &\valstd{5.98 }{0.07} &\valstd{2.25 }{0.07 } &\valstd{2.90 }{0.08 } \\
~ & w/o VE &\valstd{4.67 }{0.24 } &\valstd{3.44 }{0.09 } &\valstd{1.50 }{0.01 } &\valstd{5.93 }{0.12 } &\valstd{2.38 }{0.56 } &\valstd{2.93 }{0.51 } \\
\midrule
\multirow{7}{*}{\rotatebox{90}{Extrapolation}}
~ & \textbf{\model} &\valstdb{4.92}{0.05} &\valstdb{3.65}{0.04} &\valstdb{1.64}{0.02} &\valstd{7.02 }{0.14 } &\valstdb{2.58}{0.03} &\valstdb{3.12}{0.04} \\
\cmidrule{2-8}
~ & Random &\valstd{67.37 }{0.02 } &\valstd{20.14 }{0.03  } &\valstd{6.82 }{0.01 } &\valstd{21.70 }{0.05 } &\valstd{16.61 }{0.06 } &\valstd{10.31 }{0.02 } \\
~ & w/o TA-PLM &\valstd{5.94  }{0.23  } &\valstd{4.02  }{0.10  } &\valstd{1.70 }{0.06 } &\valstd{7.24 }{0.21 } &\valstd{3.92 }{0.33 } &\valstd{4.13 }{0.20 } \\
~ & w/o VA-PLM &\valstd{4.99 }{0.05 } &\valstd{3.68 }{0.04 } &\valstd{1.92 }{0.07 } &\valstd{8.09 }{0.20 } &\valstd{2.87 }{0.18 } &\valstd{3.25 }{0.12 } \\
~ & w/o TE &\valstd{5.04 }{0.13 } &\valstd{3.74 }{0.05 } &\valstd{1.65 }{0.02 } &\valstdb{6.94}{0.06} &\valstd{2.83 }{0.09 } &\valstd{3.41 }{0.09 } \\
~ & w/o VE &\valstd{4.94 }{0.20 } &\valstd{3.68 }{0.15 }  &\valstd{1.66 }{0.01 } &\valstd{7.02 }{0.09 } &\valstd{2.59 }{0.06 } &\valstd{3.14 }{0.05 } \\
\bottomrule
\end{tabular}
\vspace{-1mm}
\end{table*}

\vspace{-1mm}
\subsection{Few-shot and Zero-shot Learning}
\begin{figure}[tb]
  \centering
  \vspace{-0.5mm}
  \subfigure[{Adaptation by ICUType}]{
    \includegraphics[width=0.485\columnwidth]{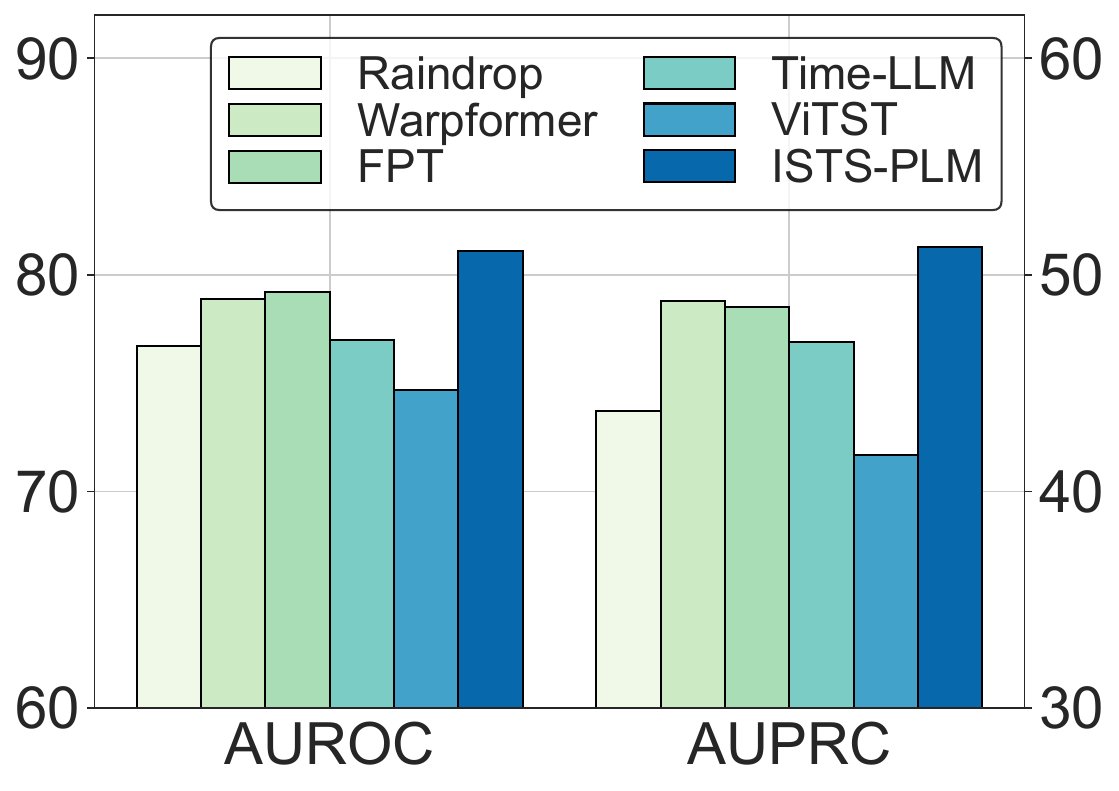}}
  \subfigure[{Adaptation by Age}]{
    \includegraphics[width=0.485\columnwidth]{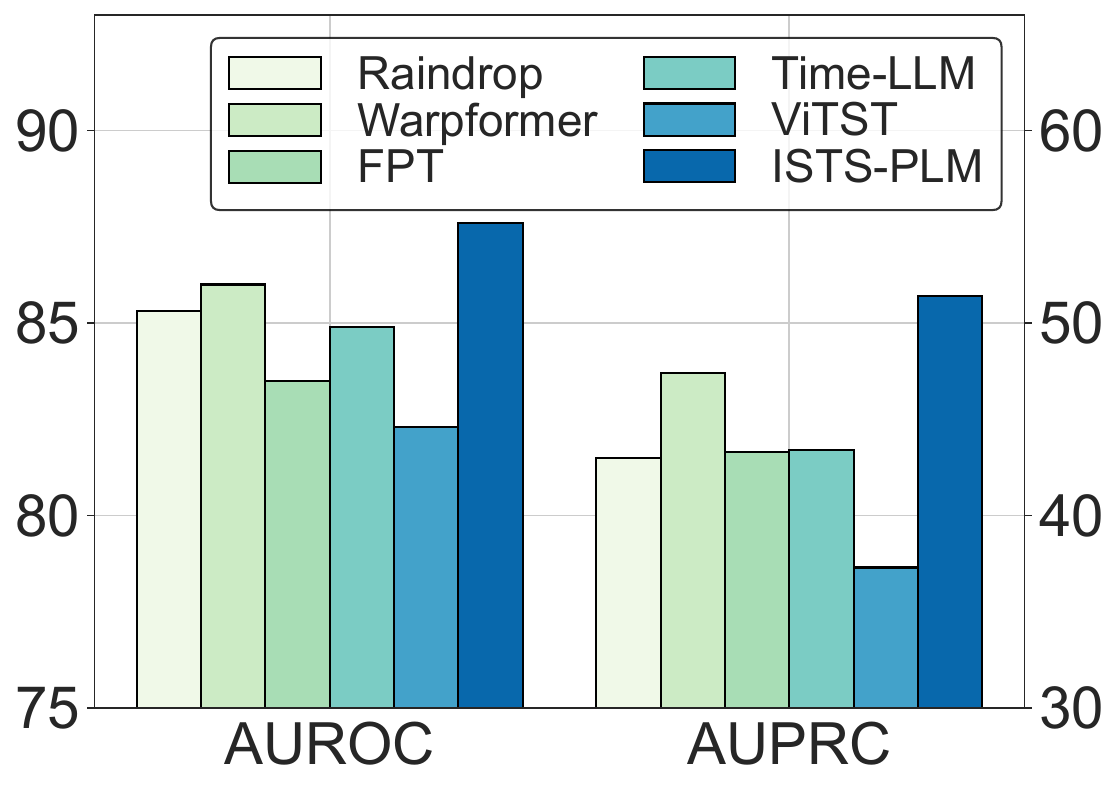}}
  \vspace{-4mm}
  \caption{Performance comparison of zero-shot adaptation.} 
  \vspace{-4mm}
  \label{fig:zero-shot}
\end{figure}
As existing ISTS works primarily focus on classification tasks, we conduct few-shot and zero-shot learning experiments on classification datasets.
For each dataset, we randomly select 10\% of the training set to assess the model's few-shot learning ability. \tabref{tab:few-shot} presents the model comparison in few-shot learning scenario, where \model consistently outperforms the other SOTA baselines. Notably, while ViTST, another cross-domain adaptation-based model for ISTS, suffers a significant performance drop, our model maintains a much more robust performance, likely due to its need to learn far fewer parameters than ViTST, as reported in \tabref{tab:model_cost}.

To evaluate the model's zero-shot adaptation ability, we divide the samples~(\ie~patients) in P12 dataset into multiple disjoint groups based on individual attributes, including ICUType~(Coronary Care Unit, Cardiac Surgery Recovery Unit, Surgical ICU, Medical ICU) and Age (Old~(>=65), Young~(<65)). This division ensures marked diversity between groups. The model is trained on some groups and tested on others.
For ICUType, we select patients belonging to Medical ICU as the test set and the others as training data. For Age, we select Young patients as the test set and Old patients as training data. The results in \figref{fig:zero-shot} showcase that \model consistently outperforms other SOTA baselines, demonstrating its robust cross-group zero-shot adaptation ability.

\subsection{Ablation Study}
We evaluate the performance of \model~and its five variants. 
(1)~\textbf{\model} represents the complete model without any ablation; 
\rev{(2)~\textbf{Random} replaces the pre-trained parameters of PLMs with randomly initialized trainable weights;}
(3)~\textbf{w/o TA-PLM} removes time-aware PLM; 
(4)~\textbf{w/o VA-PLM} removes variable-aware PLM; 
(5)~\textbf{w/o TE} removes continuous-time embeddings and directly fine-tunes the original positional embeddings in PLM;
(6)~\textbf{w/o VE} removes variable embeddings and directly fine-tunes the original positional embeddings in PLM.

The ablation results are provided in \tabref{tab:ablation_classification} and \tabref{tab:ablation_prediction}.
\rev{As can be seen, replacing the pre-trained parameters of PLMs with randomly initialized weights leads to dramatic performance collapse across all tasks and datasets, underscoring the critical role of PLMs in enhancing ISTS analysis.}
Additionally, we find removing VA-PLM generally results in a larger performance decrease in classification task, while removing TA-PLM and continuous-time embeddings overall leads to a more substantial performance drop in interpolation and extrapolation tasks. This might be attributed to classification being a higher-level task that requires greater attention to the relationships and summaries among all variables, whereas interpolation and extrapolation typically need more precise modeling of each variable's observations and their temporal flow. However, this is not absolute in the case of MIMIC. This might be because MIMIC has a large number of variables but very sparse observations, making inter-series modeling more important than intra-series modeling. Furthermore, we observe that using variable embedding in place of the original positional embeddings of PLM generally leads to better performance. This is mainly because variables are position-invariant and possess inherent semantic characteristics, while positional embeddings are sequential and lack variable-related semantics, necessitating additional effort in fine-tuning extra parameters to adapt them appropriately.

\subsection{Analysis on Distinct Representation Methods}
\label{sec:diff_reps}
\begin{figure*}[tb]
  \centering
  \vspace{-4mm}
  \subfigure[{Classification on P12}]{
    \includegraphics[width=0.53\columnwidth]{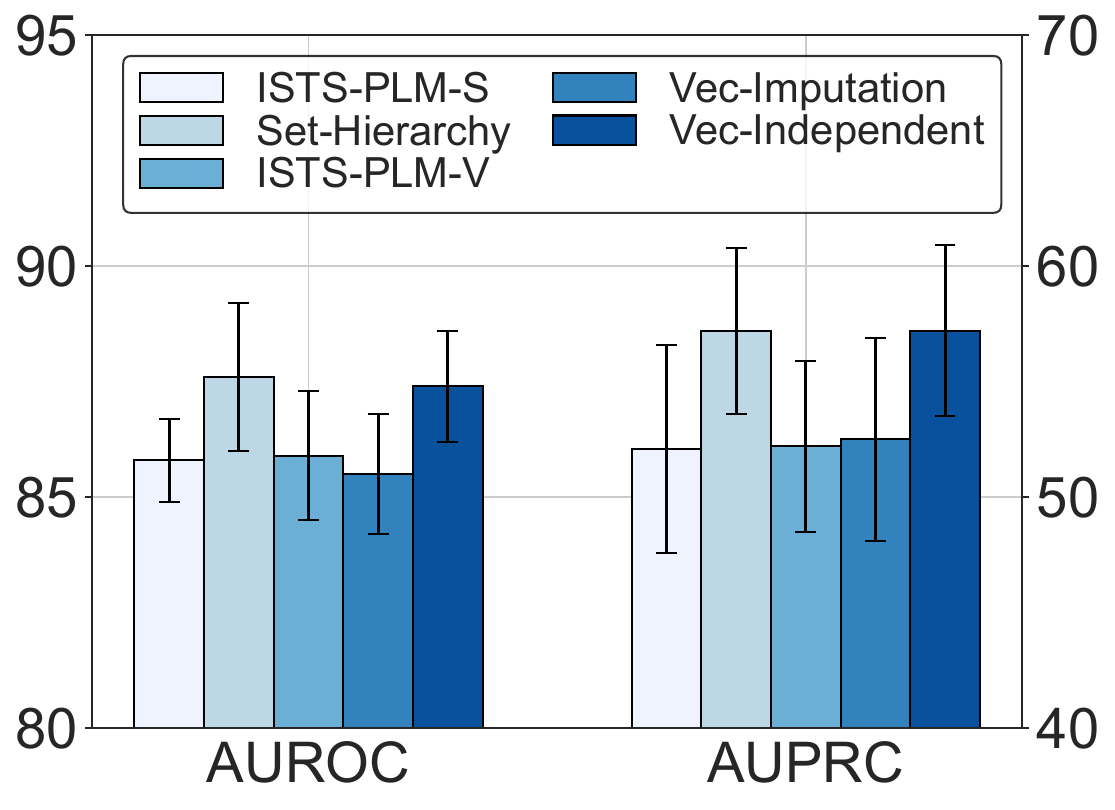}}\hspace{2mm}
  \subfigure[{Interpolation on Human Activity}]{
    \includegraphics[width=0.5\columnwidth]{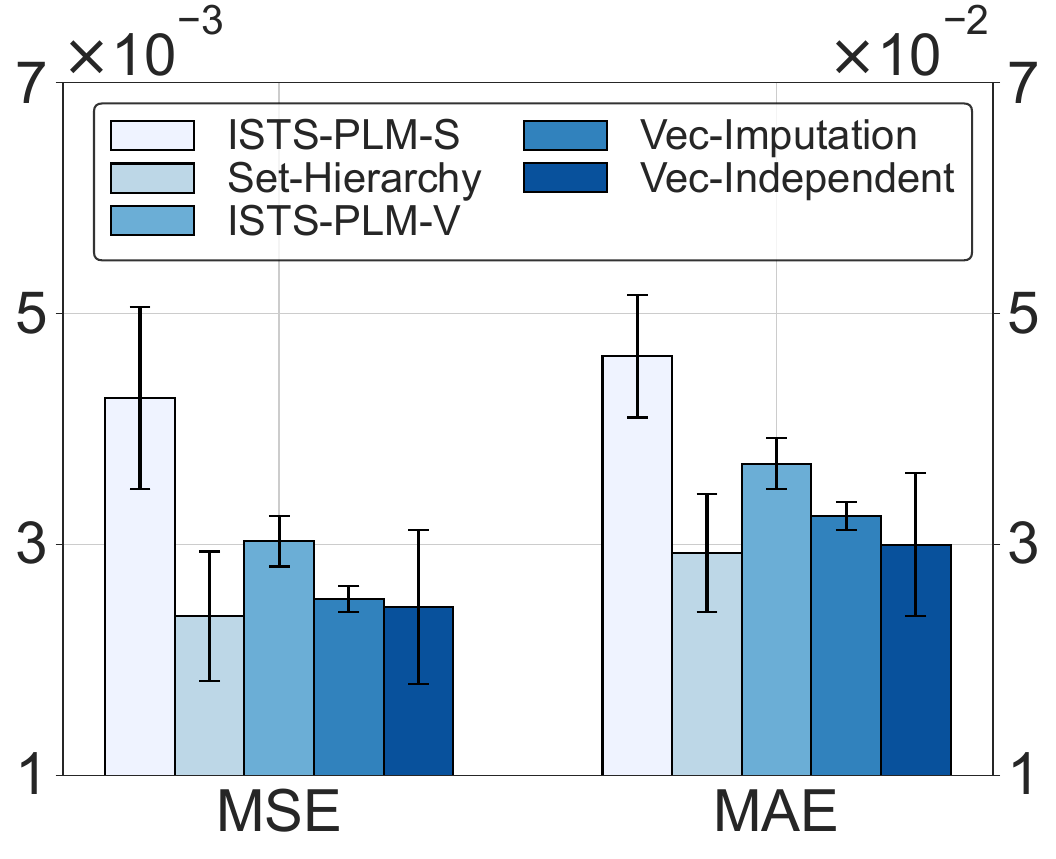}}\hspace{2mm}
  \subfigure[{Extrapolation on Human Activity}]{
    \includegraphics[width=0.5\columnwidth]{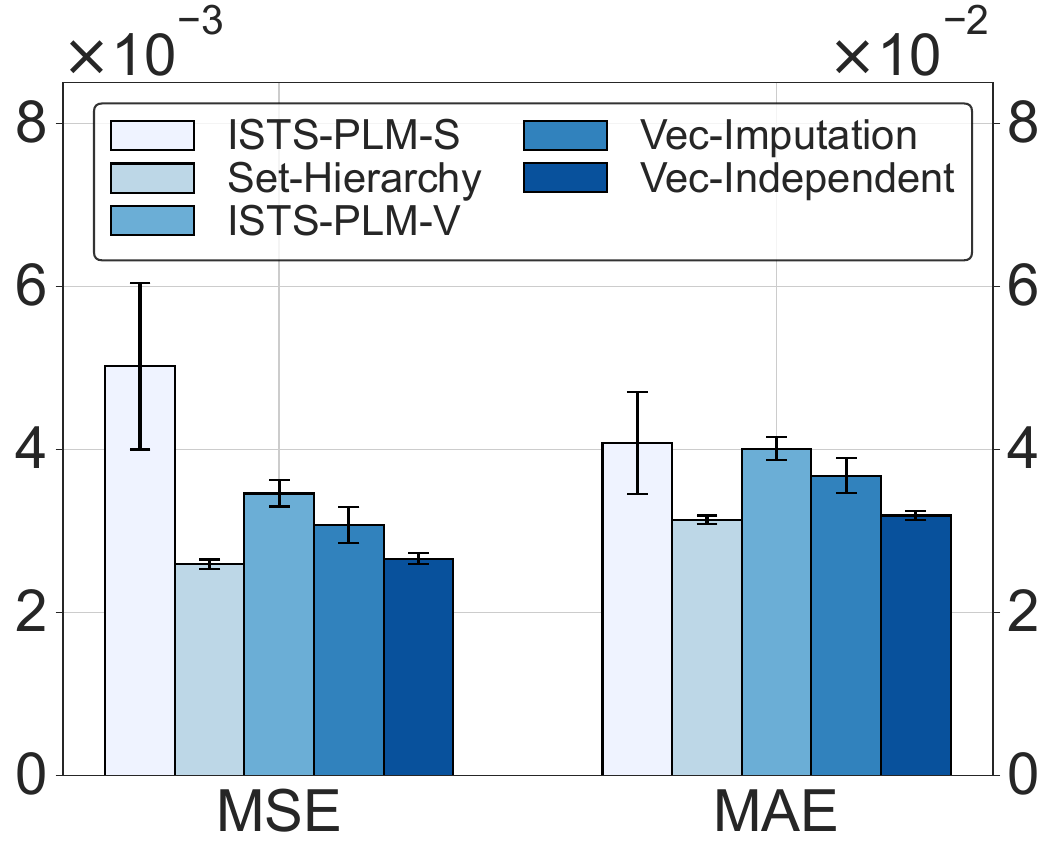}}
  \vspace{-4mm}
  \caption{Results of the variants of \model with set-based and vector-based representations.} 
  \vspace{-2.5mm}
  \label{fig:insights_reps}
\end{figure*}
This section provides a further analysis of the key failure reasons for \model when using set-based and vector-based representations. We explore several variants of \model-S and \model-V. 
For set-based representation, we examine (1)~\textbf{Set-Hierarchy}: observations are processed by PLMs in a hierarchical way, \ie~first independently modeling the observation series of each variable, then modeling the correlations between these variables. This makes it equivalent to the series-based representation.
For vector-based representation, we examine (2)~\textbf{Vec-Independent}: each variable's time series is first processed independently by the PLM, followed by PLM-based inter-variable modeling;
(3)~\textbf{Vec-Imputation}: missing values in the representation are imputed using a forward-filling strategy.

\figref{fig:insights_reps} displays the results of these variants across three ISTS analytical tasks. The findings suggest that the strategy of first modeling each variable's series independently, followed by modeling their correlations, significantly enhances the performance of PLMs in processing ISTS. 
Unlike \model-S, which models all observed tuples in a mixed time-variable manner, or \model-V, which mixes all variables' observations at each time point, this approach organizes ISTS in a more structured and coherent manner, and mitigates interference and noise from other variables, thereby simplifying the learning task for PLMs. 
\rev{In \appref{app:proof}, we attempt to further explain this finding and PLMs' universality for cross-modal learning by connecting PLMs with principal component analysis.}

Additionally, Vec-Imputation aims to assess whether the missing value issue in vector-based representation is the primary cause of failure. The results indicate that imputing missing values improves performance in interpolation and extrapolation tasks, but its impact on classification is less evident. This is probably because interpolation and extrapolation need to focus on individual data points, whereas classification relies on higher-level summarization.

\subsection{Training and Inference Cost}
\label{app:cost}
\begin{table}[tb]
\small
\centering
\caption{Comparison on training \& inference costs of classification on P12.}
\vspace{-3mm}
\label{tab:model_cost}
\setlength{\tabcolsep}{2.2mm}{\begin{tabular}{l|ccc}
\toprule
\multirow{2}{*}{\tabincell{l}{Model}} 
& \multirow{2}{*}{\tabincell{c}{Training\\ parameters}} 
& \multirow{2}{*}{\tabincell{c}{Training time\\ per step~(s)}}  
& \multirow{2}{*}{\tabincell{c}{Inference time\\ per sample~(s)}} \\
& & &\\
\midrule
GRU-D~(64) & 29K  & 0.257 & 0.045 \\
mTAND~(64) & 345K  & 0.134 & 0.010\\
Raindrop~(160) & 452K & 0.211 & 0.077 \\
Warpformer~(64) & 146K  & 0.143 & 0.017 \\
\midrule
GRU-D~(768) & 2M & 0.258 & 0.046 \\
mTAND~(768) & 20M & 0.306 & 0.069 \\
Raindrop~(772) & 11M & 1.102 & 0.966 \\
Warpformer~(768) & 16M & 0.163 & 0.021 \\
ViTST~(768) & 202M & 2.151 & 0.417 \\
\model~(768) & 127K & 0.232  & 0.043  \\
\bottomrule
\end{tabular}}
\vspace{-3mm}
\end{table}

This test is performed on a Linux server with a 20-core Intel(R) Xeon(R) Platinum 8255C CPU @ 2.50GHz and an NVIDIA Tesla V100 GPU.
\tabref{tab:model_cost} presents a comparison of training parameters, training time per update step, and inference time per sample for the classification task on P12. 
\model achieves a comparable training and inference efficiency to the recommended hidden dimensions of SOTA baselines. 
Notably, when these models are standardized to the same hidden dimension of 768, \model outperforms most of the baseline models in both training and inference efficiency. Additionally, \model requires fewer training parameters compared to most of these baselines.

\section{Conclusion}
This paper explored the potential of Pre-trained Language Models (PLMs) for Irregularly Sampled Time Series (ISTS) analysis and presented a unified PLM-based framework, ISTS-PLM, to address various ISTS analytical tasks. We investigated three methods for representing ISTS and identified a structured and coherent series-based representation that maximizes the efficacy of PLMs for ISTS modeling and analysis. 
We conducted comprehensive experiments on seven datasets spanning scientific domains of healthcare, biomechanics, and climate science. The results demonstrated that ISTS-PLM, incorporating novel time-aware and variable-aware PLMs as backbones, only fine-tuning their layer normalization parameters along with a trainable input embedding layer and a task output layer, could achieve state-of-the-art performance across various mainstream ISTS analytical tasks, such as classification, interpolation, extrapolation, few-shot and zero-shot learning scenarios compared to eighteen competitive baseline models. 


\section*{Acknowledgements}
This work was supported in part by the National Key R\&D Program of China (Grant No.2023YFF0725001), in part by the National Natural Science Foundation of China (Grant No.92370204), in part by the Guangdong Basic and Applied Basic Research Foundation (Grant No.2023B1515120057), Education Bureau of Guangzhou Municipality, and Baidu Scholarship.

\begingroup
\bibliographystyle{ACM-Reference-Format}
\def\bibfont{\fontsize{7.11}{7.11}\selectfont}
\bibliography{refs}
\endgroup


\appendix
\section{Supplementary Experiments}
\begin{table}[tb]
\small
\centering
\vspace{-3mm}
\caption{Overall \emph{Extrapolation} performance~($\operatorname{mean} \pm \operatorname{std}$) comparison on USHCN. \textbf{Bold} represents the best-performing results and \underline{underline} indicates the second-best results.}
\vspace{-3mm}
\setlength{\tabcolsep}{3mm}{\begin{tabular}{l|cc}
	\toprule
	Method & MSE$\times10^{-1}$ & MAE$\times10^{-1}$ \\ 
	\midrule
        PatchTST &\valstd{5.75}{0.01} & \valstd{3.57}{0.02}  \\
        MTGNN &\valstd{5.39}{0.05} & \valstd{3.34}{0.02}  \\
        GRU-D &\valstd{5.54}{0.38} &\valstd{3.40}{0.28} \\
        SeFT &\valstd{5.80}{0.19} &\valstd{3.70}{0.11} \\
        RainDrop &\valstd{5.78}{0.22} &\valstd{3.67}{0.17} \\
        Warpformer &\valstd{5.25}{0.05} &\valstd{3.23}{0.05}   \\
        mTAND &\valstd{5.33}{0.05} &\valstd{3.26}{0.10} \\
        Latent-ODE &\valstd{5.62}{0.03} &\valstd{3.60}{0.12} \\
        CRU &\valstd{6.09}{0.17} &\valstd{3.54}{0.18} \\
        Neural Flow &\valstd{5.35}{0.05} &\valstd{3.25}{0.05} \\
        t-PatchGNN & \valstdb{5.00}{0.04} & \valstdu{3.08}{0.04} \\
	\cmidrule{1-3}
        \model & \valstdu{5.23 }{0.07 } & \valstdb{2.99}{0.07 }  \\
	\bottomrule
\end{tabular}}
\vspace{-3mm}
\label{table:ushcn_performance}
\end{table}

\subsection{Parameter Sensitivity}
\label{app:para_sens}

\begin{figure}[tb]
  \centering
  \hspace{-2mm}
  \subfigure[{PhysioNet}]{
    \includegraphics[width=0.333\columnwidth]{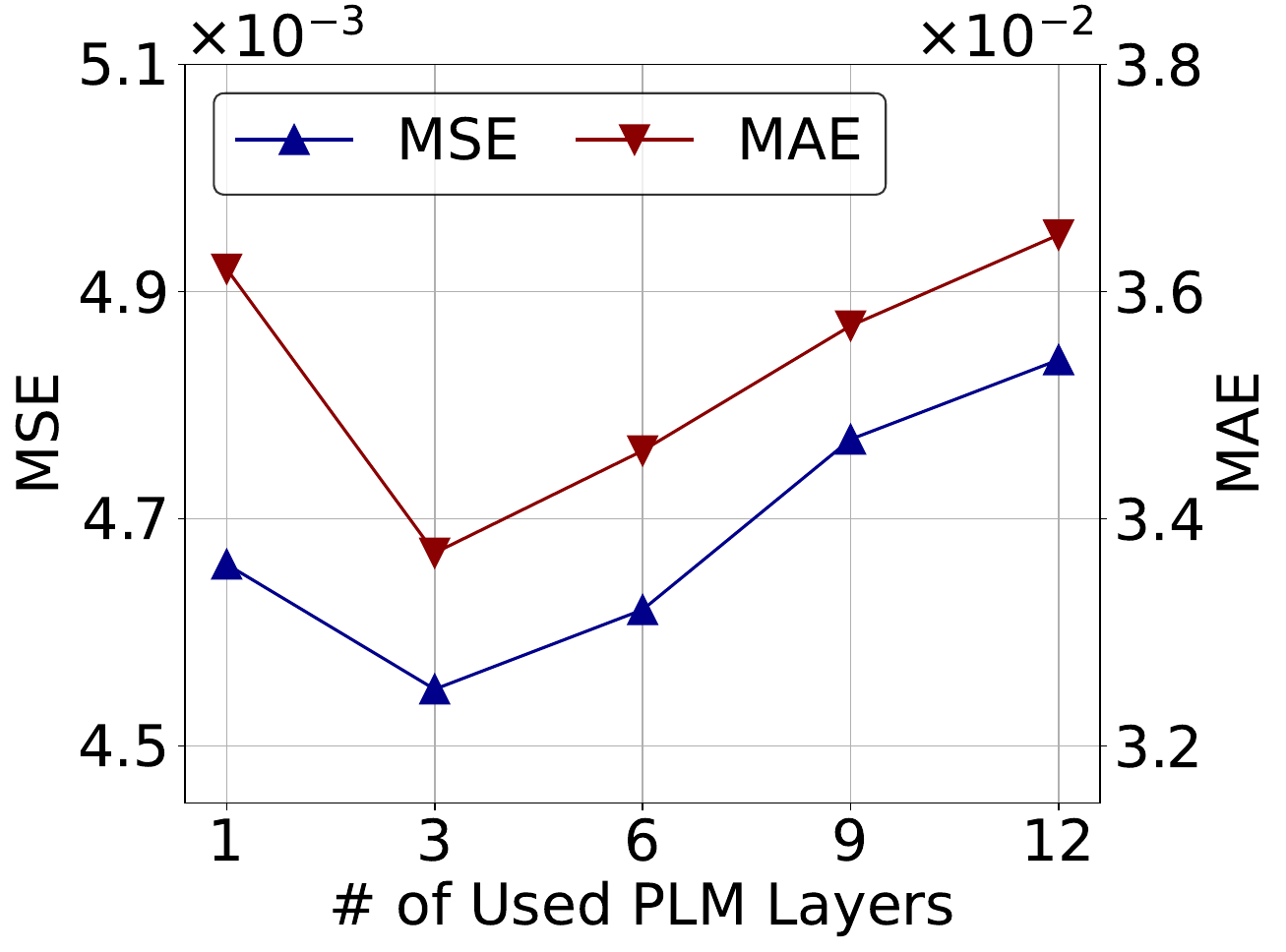}}\hspace{-1.5mm}
  \subfigure[{MIMIC}]{
    \includegraphics[width=0.333\columnwidth]{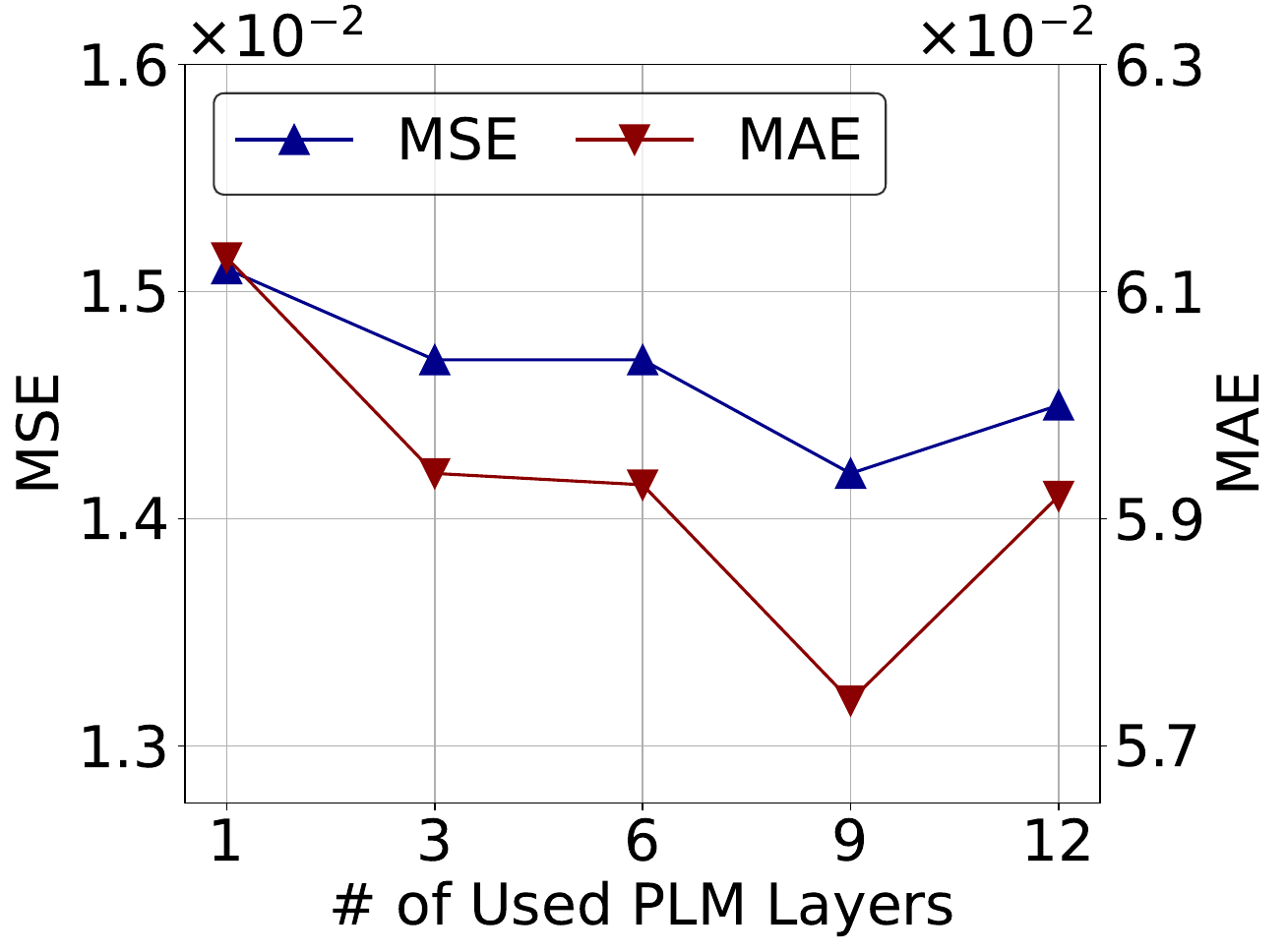}}\hspace{-1.5mm}
  \subfigure[{Human Activity}]{
    \includegraphics[width=0.333\columnwidth]{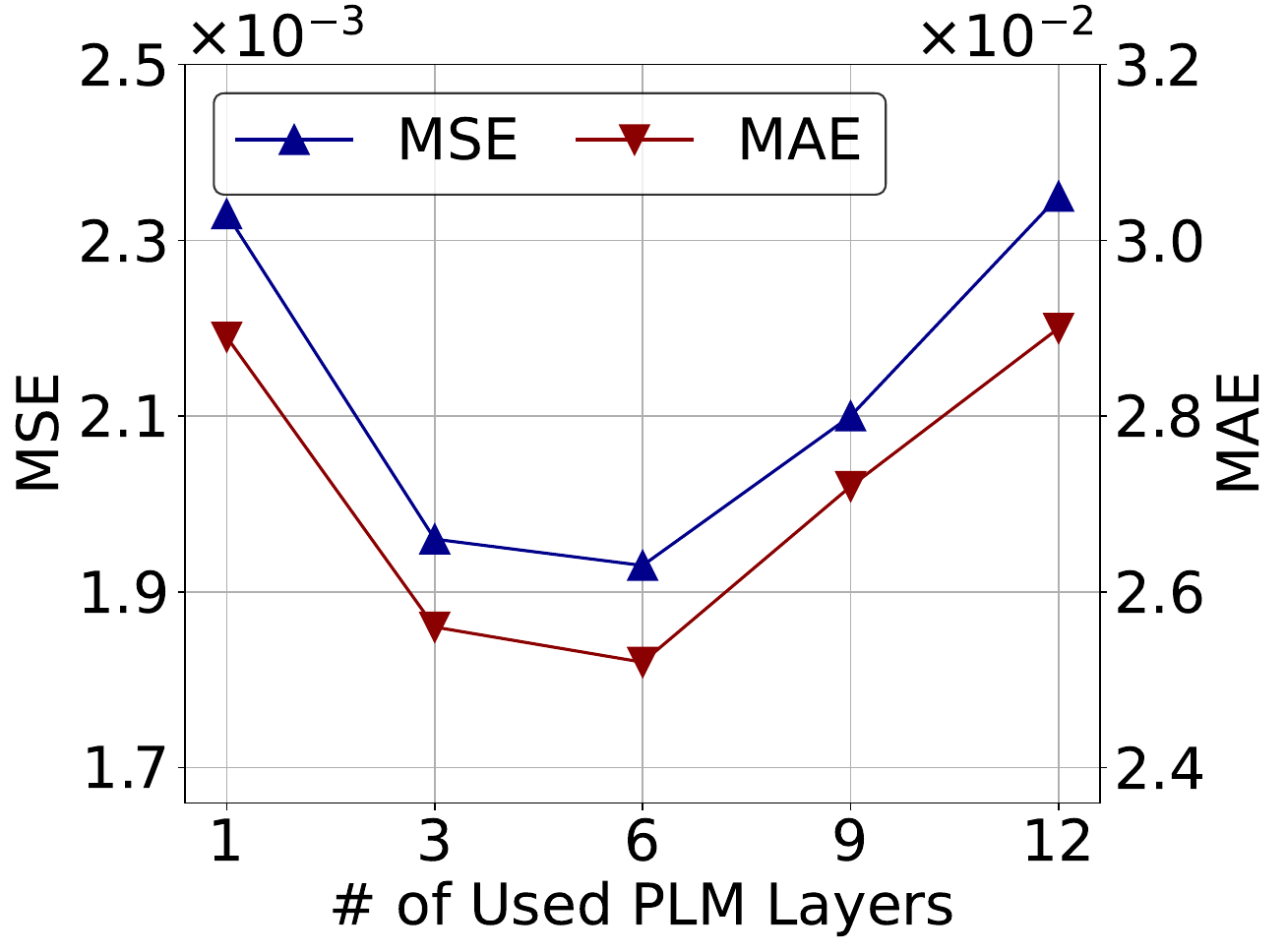}}
  \vspace{-5mm}
  \caption{Effect of different numbers of used layers in PLMs.}  
  \vspace{-4mm}
  \label{fig:param_layer}
\end{figure}
Taking the interpolation task as a representative, we study the effect of using different numbers of PLMs' layers on various datasets. As shown in \figref{fig:param_layer}, the optimal configuration for PLM's layer can vary across datasets. However, using too less~(\ie~1) or too many~(\ie~12) PLM's layers usually results in poorer performance. This is because too few layers may fail to capture the necessary complexity and dependencies within ISTS, while too many layers can complicate the process of adaptation and optimization, increasing the risk of overfitting.

\subsection{Effect of Distinct PLMs Configuration}
\label{app:diff_plms}

\rev{\tabref{tab:effect_plm_classify} and \tabref{tab:effect_plm} present the results of different PLMs backbone configurations for our model's time-aware PLM and variable-aware PLM modules. 
We find that no single configuration is universally optimal across all contexts, as their performance varies depending on the task and dataset. However, the results indicate that GPT-BERT consistently demonstrates relatively robust performance across different tasks and datasets.
This might be attributed to GPT's causal masking pre-training strategy and unidirectional, autoregressive properties, making it effective in modeling sequences where the order of data points is important, such as intra-time series dependencies. In contrast, BERT's bidirectional and contextual understanding, derived from pre-training to consider both preceding and succeeding contexts, allows it to capture complex interactions between multiple variables effectively.}

\begin{table}[tb]
\centering
\footnotesize
\vspace{-3mm}
\caption{Effect of distinct PLMs configuration for \model on \textit{Classification} task.}
\vspace{-3mm}
\label{tab:effect_plm_classify}
\setlength{\tabcolsep}{0.25mm}{
\begin{tabular}{c|cc|cc|cccc}
\toprule
& \multicolumn{2}{c|}{P12} & \multicolumn{2}{c|}{P19} & \multicolumn{4}{c}{PAM} \\ 
\cmidrule{2-9}
\multirow{-2}{*}{Config} & AUROC & AUPRC & AUROC & AUPRC & Accuracy & Precision & Recall & F1 score \\ 
\midrule
GPT-Bert   & \valstd{87.6}{1.4}  & \valstd{57.6}{3.3}  & \valstd{89.4}{2.2}  & \valstd{56.9}{5.0}  & \valstd{96.3}{0.5}  & \valstd{96.9}{1.0}  & \valstd{96.8}{0.4}  & \valstd{96.8}{0.7} \\
GPT-GPT    & \valstd{87.6}{1.4}  & \valstd{57.4}{3.4}  & \valstd{89.2}{2.4}  & \valstd{56.5}{4.5}  & \valstd{95.8}{0.5}  & \valstd{96.4}{0.4}  & \valstd{96.3}{0.5}  & \valstd{96.3}{0.4} \\
Bert-Bert  & \valstd{87.8}{1.3}  & \valstd{57.4}{3.3}  & \valstd{89.7}{1.8}  & \valstd{58.4}{2.5}  & \valstdb{96.9}{0.5} & \valstdb{97.3}{0.6} & \valstdb{97.2}{0.3} & \valstdb{97.2}{0.4} \\
Bert-GPT   & \valstdb{88.0}{1.2} & \valstdb{58.0}{2.4} & \valstdb{90.0}{2.0} & \valstdb{58.5}{3.6} & \valstd{96.4}{0.9}  & \valstd{96.8}{0.9}  & \valstd{96.7}{0.6}  & \valstd{96.7}{0.7} \\
\bottomrule
\end{tabular}
}
\vspace{-2mm}
\end{table}

\begin{table}[tb]
\centering
\footnotesize
\caption{Effect of distinct PLMs configuration for \model on \textit{Interpolation} and \textit{Extrapolation} tasks.}
\vspace{-3mm}
\label{tab:effect_plm}
\setlength{\tabcolsep}{0.1mm}{
\begin{tabular}{c|c|cc|cc|cc}
\toprule
~ & ~ & \multicolumn{2}{c|}{PhysioNet} & \multicolumn{2}{c|}{MIMIC} & \multicolumn{2}{c}{Human Activity} \\ 
\cmidrule{3-8}
\multirow{-2}{*}{Task} & \multirow{-2}{*}{Config} 
        & MSE$\times10^{-3}$ & MAE$\times10^{-2}$ 
        & MSE$\times10^{-2}$ & MAE$\times10^{-2}$ 
        & MSE$\times10^{-3}$ & MAE$\times10^{-2}$  \\ 
\midrule
\multirow{4}{*}{\rotatebox{90}{Interp.}} ~ & GPT-Bert &\valstdb{4.55}{0.08} &\valstdb{3.37}{0.02} &\valstd{1.47 }{0.01} &\valstd{5.94 }{0.01} &\valstdb{1.93}{0.01} &\valstdb{2.52}{0.01} \\
~ & GPT-GPT &\valstd{4.79 }{0.13 } &\valstd{3.46 }{0.03 }  &\valstd{1.54 }{0.01 } &\valstd{6.17 }{0.17 } &\valstd{2.05 }{0.08 } &\valstd{2.64 }{0.08 } \\
~ & Bert-Bert &\valstd{4.75 }{0.03 } &\valstd{3.56 }{0.04 }  &\valstdb{1.45}{0.01} &\valstdb{5.86}{0.08} &\valstd{1.94 }{0.01 } &\valstdb{2.52}{0.01} \\
~ & Bert-GPT &\valstd{4.76 }{0.14 } &\valstd{3.53 }{0.11 }  &\valstd{1.56 }{0.01 } &\valstd{6.27 }{0.11 } &\valstd{2.35 }{0.51 } &\valstd{2.87 }{0.42 } \\
\midrule
\multirow{4}{*}{\rotatebox{90}{Extrap.}} ~ & GPT-Bert & \valstd{4.92 }{0.05 } &\valstd{3.65 }{0.04 } &\valstdb{1.64}{0.02} &\valstd{7.02 }{0.14 } &\valstdb{2.58}{0.03} &\valstdb{3.12}{0.04} \\
~ & GPT-GPT &\valstd{4.96 }{0.05 } &\valstd{3.65 }{0.03 }  &\valstd{1.67 }{0.02 } &\valstd{6.85 }{0.07 } &\valstd{2.74 }{0.11 } &\valstd{3.20 }{0.08 } \\
~ & Bert-Bert &\valstd{4.87 }{0.13 } &\valstd{3.70 }{0.09 }  &\valstdb{1.64}{0.01} &\valstdb{6.81}{0.09} &\valstd{2.59 }{0.04 } &\valstd{3.13 }{0.04 } \\
~ & Bert-GPT &\valstdb{4.82}{0.16} &\valstdb{3.59}{0.08}  &\valstd{1.68 }{0.02 } &\valstd{6.88 }{0.06 } &\valstd{2.73 }{0.14 } &\valstd{3.21 }{0.12 } \\
\bottomrule
\end{tabular}
}
\vspace{-3mm}
\end{table}



\subsection{Dataset Details}
\label{app:datasets}


\textbf{P12/PhysioNet}~\cite{physionet}~(PhysioNet Mortality Prediction Challenge 2012\footnote{\url{https://physionet.org/content/challenge-2012/1.0.0/}}) includes 11,988/12,000 ICU patient records. Each patient is measured by 36 irregularly sampled sensors and has 5 static demographics from the first 48 hours of ICU stay. 
The goal of classification is to predict in-hospital mortality (death/survival).
For interpolation, we randomly mask 30\% timestamps' observations within ISTS and reconstruct them using the unmasked observed data.
For extrapolation, we employ the initial 24-hour period as observed data to predict subsequent 24 hours.

\textbf{P19}~\cite{reyna2019early}~(PhysioNet Sepsis Early Prediction Challenge 2019\footnote{\url{https://physionet.org/content/challenge-2019/1.0.0/}}) comprises clinical data from 38,803 patients, aiming to predict the onset of sepsis~(a binary label) within the next 6 hours. Each patient is monitored using 34 irregularly sampled sensors, including 8 vital signs and 26 laboratory values, alongside 6 demographic features. This binary classification task is highly imbalanced, with only around 4\% positive samples.


\textbf{PAM}~\cite{reiss2012introducing}~(PAMAP2 Physical Activity Monitoring\footnote{\url{https://archive.ics.uci.edu/ml/datasets/pamap2+physical+activity+monitoring}}) 
contains sensor data from 8 subjects (1 excluded for short recordings), each performing 18 physical activities with 3 wearable IMUs. We keep 8 activities with more than 500 samples, yielding an 8-class classification task with total 5,333 samples (600 time steps each) in the final dataset. To simulate irregularity, 60\% of observations are randomly removed. No static features are included, and the classes are approximately balanced.

\textbf{MIMIC}~\cite{mimic3}~(Medical Information Mart for Intensive Care\footnote{\url{https://mimic.mit.edu/}}) is a comprehensive clinical database that includes electronic health records in critical care. This dataset contains 23,457 ISTS samples for different patients, covering the first 48 hours following their admission, and each patient comprises 96 variables. 
For interpolation, we randomly mask 30\% timestamps' observations within ISTS and reconstruct them using the unmasked observed data.
For extrapolation, we employ the initial 24-hour period as observed data to predict subsequent 24 hours.

\textbf{Human Activity}~\cite{misc_localization_data_for_person_activity_196}~(Localization Data for Person Activity\footnote{\url{https://archive.ics.uci.edu/dataset/196/localization+data+for+person+activity}}) includes 12 variables derived from irregular 3D positional data collected by four sensors from five individuals performing various activities such as walking, sitting, lying down, and standing. 
We chunk the original time series into 5,400 ISTS samples, each spanning 4,000 milliseconds. 
For interpolation, we randomly mask 30\% timestamps' observations within ISTS and reconstruct them using the unmasked observed data.
For extrapolation, we utilize the initial 3,000 milliseconds as observed data to predict the queried positional values of sensors in the subsequent 1,000 milliseconds. 

\textbf{USHCN}~\cite{USHCN}~(The United States Historical Climatology Network\footnote{\url{https://www.osti.gov/biblio/1394920}})
comprises daily measurements of 5 climate variables~(snowfall, snow depth, precipitation, maximum/minimum temperature) collected from 1,114 U.S. meteorological stations. We follow the pre-processing of~\cite{de2019gru} to retain 5\% of observations from each station during the years 1996–1999. We chunk the dataset into 26,736 ISTS samples, each comprising 25 consecutive months of climate data. 
For extrapolation, we use observed data of preceding 24 months to predict climate variables during subsequent month. 

\subsection{Baseline Details}
\label{app:baseline}
We incorporate eighteen baselines into our experiments for a fair comparison. The settings and results of classification and extrapolation tasks entirely refer to ViTST~\cite{li2023time} and t-PatchGNN~\cite{zhangirregular2024}, respectively. 
In terms of the interpolation task, we meticulously tune the key hyperparameters of the baseline models around their recommended settings. 
We standardize the hidden dimensions to 64 for Physionet and MIMIC, and to 32 for Human Activity and USHCN, and use Adam optimizer for training.
To adapt these baseline models to interpolation task, we replace their original output layer with the interpolation prediction layer instantiated by a three-layer MLP.
As the setups of most baselines for classification and extrapolation have been detailed in ViTST~\cite{li2023time} and t-PatchGNN~\cite{zhangirregular2024}, respectively, we primarily present their interpolation settings and provide links to the implementations we used in the footnotes.


\textbf{GRU-D}\footnote{\url{https://github.com/zhiyongc/GRU-D}}~\cite{che2018recurrent} handles missing values and irregular timestamps by introducing trainable decay terms into a GRU architecture. We use a learning rate of $1\times10^{-3}$.

\textbf{SeFT}\footnote{\url{https://github.com/mims-harvard/Raindrop}}~\cite{horn2020set} transforms time series into a set encoding and utilizes set functions to model them. In our experiment, we configure it with 2 layers and a learning rate of $1\times10^{-3}$. 

\textbf{Raindrop}\footnote{\url{https://github.com/mims-harvard/Raindrop}}~\cite{zhang2021graph} employs Graph Neural Networks~(GNNs) and temporal self-attention to estimate missing values by leveraging inter-sensor correlations. We configure it with 2 Transformer layers and heads, observation dim of 4, and a learning rate of $1\times10^{-3}$. 

\textbf{Warpformer}\footnote{\url{https://github.com/imJiawen/Warpformer}}~\cite{warpformer2023} is a Transformer-based model that aligns ISTS via a warping module and enhances learning through a customized attention mechanism.  
In our experiment, we set the warp number to 0-0.2-1, with 1 head and 3 layers for classification, and 2 layers for interpolation. The learning rate is set to $1\times10^{-3}$. 

\textbf{mTAND}\footnote{\url{https://github.com/reml-lab/mTAN}}~\cite{shukla2020multi} is a model for ISTS classification and interpolation, which learns embeddings for numerical values tied to continuous time steps and derives fixed-length representations for variable-length sequential data using an attention mechanism.
In our experiments, we set k-iwae to 5, std to 0.01, the number of reference points to 64, and the learning rate to $1\times10^{-3}$. 

\textbf{Latent-ODE}\footnote{\url{https://github.com/YuliaRubanova/latent_ode}}~\cite{rubanova2019latent} is an ODE-based model that enhances RNNs by incorporating continuous-time hidden state dynamics specified by neural ODEs. In our experiment, we configure it with 3 rec-layers and gen-layers for PhysioNet, and 1 for MIMIC and Human Activity. The learning rate is set to $1\times10^{-3}$. 

\textbf{Neural Flow}\footnote{\url{https://github.com/mbilos/neural-flows-experiments}}~\cite{rubanova2019latent}~\cite{bilovs2021neural}  leverages flow-based generative models to parameterize the solution curves of ODEs. For MIMIC, we use 4 flow layers and a GRU-based flow; for others, 2 flow layers with a coupling flow model. We apply ReLU or Tanh activations and set the learning rate to $1\times10^{-3}$. 

\textbf{CRU}\footnote{\url{https://github.com/boschresearch/Continuous-Recurrent-Units}}~\cite{schirmer2022modeling} integrates Kalman filtering and ODE modeling via an encoder-decoder architecture for continuous state estimation. Time scaling is set to 0.3 for Activity and 0.2 otherwise. Activation choices include squared variance (encoder) and exponential (decoder), with ReLU for transitions and learning rate $1\times10^{-3}$. 


\textbf{t-PatchGNN}\footnote{\url{https://github.com/usail-hkust/t-PatchGNN}}~\cite{zhangirregular2024} is a SOTA model for ISTS extrapolation that uses a transformable patching approach to address both irregularity and asynchrony, followed by time-adaptive GNNs to model dynamic inter-series correlations. For the interpolation task, the patch window size for PhysioNet and MIMIC is set to 8 hours, and 300 milliseconds for Human Activity. The learning rate is $1\times10^{-3}$. 

\textbf{FPT}\footnote{\url{https://github.com/DAMO-DI-ML/NeurIPS2023-One-Fits-All}}~\cite{zhou2023one} introduces a frozen PLM to address RSTS analytical tasks. For interpolation and extrapolation, we use the following settings: the representation method of input is the same as \model~(\ie~series-based). We use 6 PLM layers, patch size 48, stride 24, and GeLU output activation. The learning rate is $5\times10^{-4}$. 

\textbf{Time-LLM}\footnote{\url{https://github.com/KimMeen/Time-LLM}}~\cite{jin2024time} propose a framework to reprogram an existing LLM to perform RSTS forecasting. For interpolation and extrapolation, we use the following settings: the representation method of input is the same as \model~(\ie~series-based).
The size of the patch is 48 and the length of the stride is 24 to adapt the representation of ISTS. The learning rate is $1\times10^{-4}$. 

\section{Discussion on PLMs' Universality in Cross-modal Learning}
\label{app:proof}
Prior work~\cite{zhou2023one} has theoretically demonstrated the self-attention mechanism in PLMs functions similarly to Principal Component Analysis (PCA). This reveals the universality of PLMs in bridging the modality gap between language and time series and may explain why the series-based representation outperforms both set-based and vector-based methods. 
From this perspective, the set-based method applies ``PCA'' to the mixed time-variable dimensions, which disturbs the structure of ISTS and impairs the effectiveness of ``PCA''. The vector-based method only applies ``PCA'' to the intra-series dimension, neglecting the inter-series dimension. In contrast, the series-based method organizes ISTS in a more structured and coherent manner and applies ``PCA'' on both intra-series (via time-aware PLMs) and inter-series dimensions (via variable-aware PLMs), thereby more effectively harnessing the power of PLMs for ISTS.  

\end{document}